\documentclass[11pt]{article}

\usepackage[preprint]{acl}
\usepackage{times}
\usepackage{latexsym}
\usepackage[most]{tcolorbox}
\usepackage{xcolor}
\definecolor{paperred}{RGB}{182, 48, 28}
\definecolor{paperyellow}{RGB}{196, 135, 20}
\definecolor{papergreen}{RGB}{46, 125, 50}

\usepackage[T1]{fontenc}

\usepackage[utf8]{inputenc}

\usepackage{microtype}

\usepackage{inconsolata}

\usepackage{graphicx}

\usepackage{algorithm}
\usepackage{algpseudocode}
\usepackage{amsfonts}
\usepackage{amsmath}
\usepackage{amssymb}
\usepackage{booktabs}
\usepackage{colortbl}
\usepackage{makecell}
\usepackage{multirow}
\usepackage{subcaption}
\usepackage{xcolor}

\title{Interpretable Text Embeddings Based on Hierarchical Questions}

\author{
  Kohei Oda\textsuperscript{1}{\qquad}Po-Min Chuang\textsuperscript{2}{\qquad}Kiyoaki Shirai\textsuperscript{1}{\qquad}Natthawut Kertkeidkachorn\textsuperscript{1} \\
  $^{1}$Japan Advanced Institute of Science and Technology \\
  $^{2}$Toshiba Corporation \\
  \textsuperscript{1}\texttt{\{s2420017, kshirai, natt\}@jaist.ac.jp} \\
  \textsuperscript{2}\texttt{pomin.chuang.x51@mail.toshiba}
}

\begin{document}
\maketitle
\begin{abstract}
QA-based text embeddings represent texts as vectors consisting of 1/0 answers to yes/no questions.
Although these embeddings provide detailed information for interpretation and achieve high task performance, they often require humans to examine a large number of questions.
Furthermore, the relationship between the number of questions and the appropriate level of question granularity has not been sufficiently studied.
To address these challenges, we propose a method for constructing hierarchical questions for text embeddings.
As the hierarchy deepens, questions become more specific and their number increases.
During interpretation, we reduce the number of questions that humans need to examine by replacing questions at deeper levels with aligned questions at shallower levels.
The question hierarchy is constructed through recursive clustering, contrastive prompting, and question alignment.
Experimental results on general-purpose tasks show that our method improves interpretability while maintaining task performance.
It is also found that, when only a limited number of questions are available, questions at shallower levels tend to achieve higher task performance than those at deeper levels, even with the same number of questions.\footnote{Our code is publicly available at \url{https://github.com/iehok/HierarchicalQA_Embeddings}}
\end{abstract}

\section{Introduction}
Text embeddings, which represent sentences, paragraphs, and documents as vectors, are fundamental to a wide range of natural language processing (NLP) applications, such as information retrieval~\citep{thakur2021beir} and enhancing the generative capabilities of large language models (LLMs)~\citep{lewis2021retrievalaugmentedgenerationknowledgeintensivenlp}.
For this reason, methods for obtaining high-quality text embeddings have been extensively studied~\citep{reimers-gurevych-2019-sentence, gao-etal-2021-simcse, morris2025contextual}.
However, these embeddings remain difficult to interpret, making it challenging for humans to understand which semantic aspects contribute to the similarity or dissimilarity between two texts.
This lack of interpretability is particularly critical in high-stakes domains, such as legal and medical applications, where transparent NLP systems are highly desirable~\citep{mcinerney-etal-2023-chill}.

\begin{figure}[!t]
    \centering
    \includegraphics[width=1.0\linewidth]{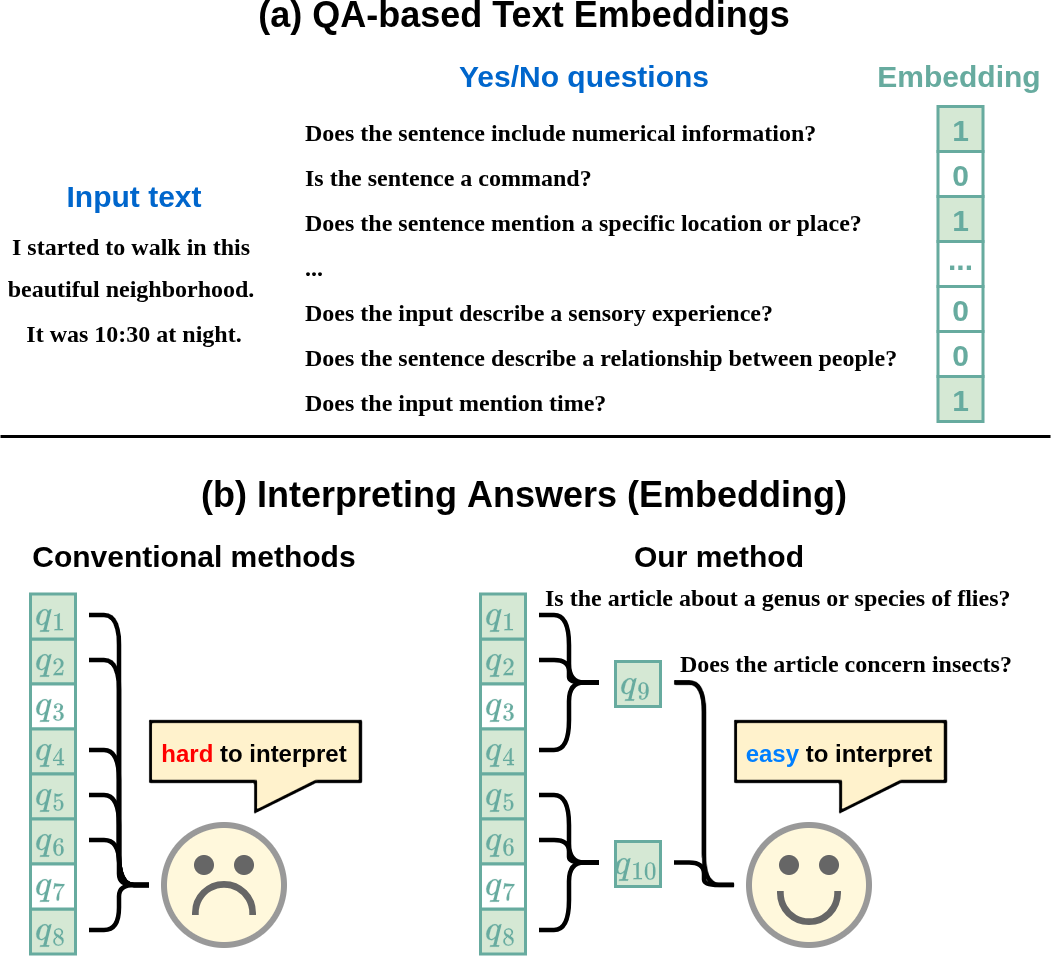}
    \caption{Overview of QA-based text embeddings and interpreting each embedding.}
    \label{fig:overview}
\end{figure}

To develop interpretable text embeddings, various methods have been proposed~\citep{wang-etal-2025-ldir, anand-etal-2026-ibert, shin2026aligning}.
In particular, QA-based text embeddings~\citep{benara2024crafting, sun2025a, tang2026askingrightquestionsontologygrounded} shown in Figure~\ref{fig:overview}~(a) are highly interpretable because each dimension of an embedding corresponds to a yes/no question and represents its binary (1/0) answer.
While conventional QA-based text embeddings represent texts using answers to relatively detailed questions, they tend to require a large number of questions for human interpretation.
Furthermore, it remains unclear how many questions are needed and how specific they should be for general-purpose tasks.

This study organizes questions used for text embeddings into a hierarchical structure, where more specific questions are allocated at the lower levels.
This approach has two advantages.
First, it can flexibly change the number of dimensions of embeddings, i.e., the number of questions.
It is preferable, since the optimal number of dimensions may differ across downstream tasks for which the text embeddings are applied.
Second, it can alleviate the burden on humans to interpret text embeddings.
Instead of reading all questions, a human can refer to a small number of general questions, as illustrated in Figure~\ref{fig:overview}~(b), where the specific questions $q_1$, $q_2$, and $q_4$ are summarized by the general question $q_9$.

\begin{figure}[!t]
    \centering
    \includegraphics[width=1.0\linewidth]{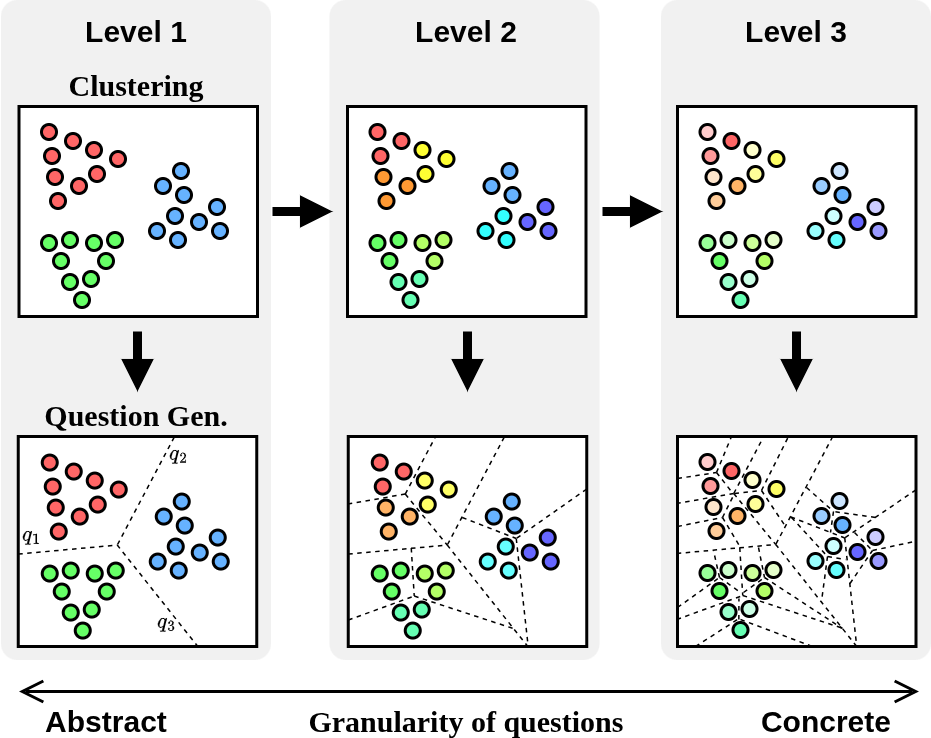}
    \caption{Overview of question generation.}
    \label{fig:method}
\end{figure}

To generate questions of varying granularity, as shown in Figure~\ref{fig:method}, we recursively apply the question-generation framework proposed by \citet{sun2025a}, which combines clustering of large-scale text corpora with contrastive prompting.
Contrastive prompting generates highly discriminative questions, and recursive clustering is expected to result in questions at shallower levels summarizing questions at deeper levels.
The alignment between a question at a given level and a question at the preceding level is determined based on their entailment relationship.

Experiments on three tasks, semantic textual similarity (STS)~\citep{agirre-etal-2012-semeval}, text clustering~\citep{Aggarwal2012}, and information retrieval~\citep{thakur2021beir} show that our method substantially improves the Pareto front between embedding performance and cognitive load.
Our analysis also shows that questions become more specific at deeper levels of the hierarchy, and that aligned questions have an entailment relation with high probability.
We further find that, when only a limited number of questions are available, questions at shallower levels produce better text embeddings than questions randomly selected at deeper levels.

\section{QA-based Text Embeddings}
\label{sec:qa_based_text_embeddings}
QA-based text embeddings~\citep{benara2024crafting, sun2025a, tang2026askingrightquestionsontologygrounded} represent a text through its responses to a set of natural-language questions.
Each dimension of the embedding corresponds to a yes/no question, with a value of $1$ or $0$ indicating a ``yes'' or ``no'' response, respectively.
For instance, the text ``A study shows that exercise is good for health.'' may be represented as a vector of $[1, 0, 1]$ for the questions ``Is it about health?'', ``Is it about politics?'', and ``Does it describe research?''.
Thus, QA-based embeddings allow texts to be compared in semantic space while preserving the interpretability of individual dimensions.

Let $\mathcal{Q} = \{q_1, q_2, \ldots, q_m\}$ be a set of yes/no questions.
The QA-based embedding $\mathbf{e}(x)$ of an input text $x$ is expressed as follows:
\begin{equation}
    \mathbf{e}(x) = [a_1(x), a_2(x), \ldots, a_m(x)],
\end{equation}
where $a_i(x)$ denotes the binary answer to question $q_i$ for $x$.
The model $f_i(x)$ predicts the probability that the answer to question $q_i$ for sentence $x$ is ``yes'', and $a_i(x)$ is determined as follows,
\begin{equation}
    a_i(x) =
    \begin{cases}
        1\ (\text{yes})  & \text{if $f_i(x) > \tau$} \\
        0\ (\text{no})   & \text{otherwise}
    \end{cases},
\end{equation}
where $\tau$ is a threshold.
The model $f_i(x)$ is trained using a collection of sample texts and LLM-generated answers to those samples.

Since questions directly affect the quality of text embeddings, the method to create questions is a critical element in QA-based text embedding.
\citet{benara2024crafting} proposed QA-Emb, a method uses LLMs to generate yes/no questions with task-specific prompts containing examples.
\citet{sun2025a} further proposed CQG-MBQA, a framework for generating more discriminative questions for QA-based text embeddings.
It first groups texts into clusters based on the semantic similarity and then uses LLMs to generate yes/no questions that distinguish between the clusters.
Since this framework can generate interpretable text embeddings comparable to those produced by black-box text embedding methods, we build our question generation method based on this approach.

\section{Proposed Method}
Our proposed method consists of three processes: question generation, question alignment, and model training.
In the question generation process, we combine recursive clustering with the question-generation method proposed by \citet{sun2025a} to generate highly discriminative questions at various levels of granularity~(\S~\ref{subsec:q_gen}).
In question alignment, each question at a given hierachical level is matched with a question at the superior level with which it has an entailment relationship~(\S~\ref{subsec:q_map}).
In the model training process, for each generated question, we train a model that predicts the probability to answer ``yes'' for a given sentence~(\S~\ref{subsec:train_qa_models}).

\subsection{Question Generation}
\label{subsec:q_gen}
Question generation process consists of three steps: recursive clustering, contrastive question generation, and question filtering.
A simplified overview is shown in Algorithm~\ref{alg:q_gen}, and each step is explained in detail below.

\begin{algorithm}[!t]
\caption{Question Generation}
\label{alg:q_gen}
\begin{algorithmic}[1]
\State \textbf{\# Step1. Recursive Clustering}
\State $\mathcal{S} \gets \{s_1, s_2, ...\}$
\State $\mathbf{C}^1 \gets \mathrm{kmeans}(\mathcal{S}, k_\text{first})$
\For{$l \in \{2, 3, ..., l_\text{max}\}$}
    \State $\mathbf{C}^l \gets \bigcup_{\mathcal{C}^{l-1}_i \in \mathbf{C}^{l-1}}\mathrm{kmeans}(\mathcal{C}^{l-1}_i, k)$
\EndFor
\medskip
\State \textbf{\# Step2. Contrastive Question Generation}
\For{$l \in \{1, 2, ..., l_\text{max}\}$}
    \State $\mathbf{\tilde{R}}^l \gets \bigcup_{\mathcal{C}^l_i \in \mathbf{C}^l}{\{ \, \mathcal{\tilde{R}}^l_i \,|\, \mathcal{\tilde{R}}^l_i \overset{\text{def}}{=} \mathrm{cqg}(\mathcal{C}^l_i, \mathbf{C}^l)\} \, }$
    \State $\mathcal{\tilde{Q}}^l \gets \bigcup_{\mathcal{\tilde{R}}^l_i \in \mathbf{\tilde{R}}^l} \; \mathcal{\tilde{R}}^l_i$
\EndFor
\medskip
\State \textbf{\# Step3. Question Filtering}
\For{$l \in \{1, 2, ..., l_\text{max}\}$}
    \State $\mathbf{R}^l \gets \mathrm{filtering}(\mathbf{\tilde{R}}^l)$
    \State $\mathcal{Q}^l \gets \bigcup_{\mathcal{R}^l_i \in \mathbf{R}^l}{\mathcal{R}^l_i}$
\EndFor
\end{algorithmic}
\end{algorithm}

\paragraph{Step1. Recursive Clustering}
We first use the MEDI2 dataset\footnote{\url{https://huggingface.co/datasets/GritLM/MEDI2}}~\citep{muennighoff2025generative}, which contains approximately 6.8 million texts, as a large-scale corpus $\mathcal{S}$.
The recursive $k$-means clustering is applied to $\mathcal{S}$, where each text is encoded using an existing black-box text embedding model.
At the first level, $\mathcal{S}$ is grouped into $k_\text{first}$ clusters $\mathbf{C}^1$.
At subsequent levels, texts in each cluster from the previous level $\mathcal{C}^{l-1}_i$ are grouped into $k$ clusters, and $\mathbf{C}^l$ is obtained by merging them.
Thus, the number of clusters increases by a factor of $k$ at each level, i.e., $|\mathbf{C}^l| = k |\mathbf{C}^{l-1}|$.
The parameter $k_\text{first}$ is set larger than $k$ to make a relatively large number of clusters at the top level.

\paragraph{Step2. Contrastive Question Generation}
For each cluster $\mathcal{C}^l_i$ at the level $l$, questions are generated by the contrastive question generation (CQG) framework proposed by \citet{sun2025a}.
First, positive and negative samples are collected.
The positive samples are $n_p$ texts randomly sampled from $\mathcal{C}^l_i$.
The negative samples include \textit{hard negatives} and \textit{easy negatives}.
For the hard negatives, $n_h$ texts are randomly sampled from the $n_c$ clusters in $\mathbf{C}^l$ that are closest to $\mathcal{C}^l_i$.
For the easy negatives, $n_e$ texts are randomly sampled from the remaining clusters.
The sampled positive and negative texts are then included in a prompt that instructs the LLM to generate 10 questions that yield ``yes'' for the positive samples and ``no'' for the negative samples.
The prompt is provided in Appendix~\ref{apd:prompts}.
As a result of this step, 10 times as many questions are generated as there are clusters.
In Algorithm~\ref{alg:q_gen}, $\mathbf{\tilde{R}}^l$ is a collection of question sets, where each question set is generated from a single cluster, while $\mathcal{\tilde{Q}}^l$ is a set of all questions at level $l$.

\paragraph{Step3. Filtering Questions}
Many of the generated questions are semantically similar, resulting in redundancy that increases the computational cost of constructing QA-based text embeddings.
Therefore, we apply a filtering process to the generated questions following a probing mechanism proposed by \citet{sun2025a}.
First, we calculate the quality for each question $q \in \mathcal{\tilde{Q}}^l$ at the level $l$.
As in CQG, we first randomly sample $p_p$ positive probes (samples) from the cluster $\mathcal{C}^l_i$ where $q$ is generated, $p_h$ hard negative probes from $n_c$ clusters close to $\mathcal{C}^l_i$, and $p_e$ easy negative probes from other clusters.
After that, the answers to $q$ are generated by a LLM for each probe, and the quality of $q$ is calculated as follows:
\begin{align}
    \mathrm{quality}(q) &= \frac{\text{\# ``yes'' for positive probes}}{p_p} \nonumber \\
                        &- \frac{\text{\# ``yes'' for negative probes}}{p_h + p_e}.
\end{align}
Afterward, all questions are encoded using a black-box text embedding method, and questions whose cosine similarity exceeds $\theta$ are removed.
In this filtering process, it is ensured that higher-quality questions are prioritized and retained.
The detailed algorithm for this filtering process is described in Appendix~\ref{apd:filtering}.

\subsection{Question Alignment}
\label{subsec:q_map}

\begin{algorithm}[!t]
\caption{Question Alignment}
\label{alg:q_map}
\begin{algorithmic}[1]
\State $T_\text{question}$: $l_\text{max} \times |\mathcal{Q}^{l_\text{max}}|$ matrix
\For{$l \in \{2, 3, ..., l_\text{max}\}$}
    \For{$\mathcal{R}^l_i \in \mathbf{R}^l$}
        \State $j \gets T_\text{cluster}[l][i]$
        \For{$q^l \in \mathcal{R}^l_i$}
            \State $\tilde{q}^{l-1} \gets$
            \State $\mathrm{argmax}_{q^{l-1} \in \mathcal{R}^{l-1}_j} \mathrm{entail}(q^l, q^{l-1})$
            \State $a \gets$ index of $q^l$ in $\mathcal{Q}^l$
            \State $b \gets$ index of $\tilde{q}^{l-1}$ in $\mathcal{Q}^{l-1}$
            \State $T_\text{question}[l][a] \gets b$
        \EndFor
    \EndFor
\EndFor
\end{algorithmic}
\end{algorithm}

We align questions at level $l$ to the most related question at the superior level, called the ``parent question'' hereafter.
The alignment table $T_\text{question}[l][a]$ is defined as an index of the parent question of the $a$-th question in $\mathcal{Q}^l$, and computed by Algorithm~\ref{alg:q_map}.
Let $T_\text{cluster}$ denote the precomputed alignment table that records an index of the parent cluster.
For each question cluster $\mathcal{R}^l_i$ at level $l$, an index of the parent cluster ($j$) is obtained using $T_\text{cluster}$.
Next, the parent question $\tilde{q}^{l-1}$ is chosen from the questions $q^{l-1}$ in $\mathcal{R}^{l-1}_j$.
Specifically, $q^l$ and $q^{l-1}$ are converted into declarative sentences $d^l$ and $d^{l-1}$, respectively, and a natural language inference (NLI) model predicts the probability that $d^l$ entails $d^{l-1}$.
This score represents the validity of replacing $q^l$ with $q^{l-1}$ when a human interprets text embeddings.
Then, the question with the highest entailment probability is selected, and its index in $\mathcal{Q}^{l-1}$ is recorded in $T_\text{question}$.

\subsection{Model Training}
\label{subsec:train_qa_models}
The QA model $f_i(x)$ described in Section~\ref{sec:qa_based_text_embeddings} is trained based on the multi-task binary question answering (MBQA) framework proposed by \citet{sun2025a}.
For each question $q$ at the level $l$, the training data for $q$ is created as the following procedure.
First, 50 texts from the cluster $\mathcal{C}^l_i$ where question $q$ is generated, 30 texts from the $n_c$ clusters closest to $\mathcal{C}^l_i$ within $\mathbf{C}^l$, and 20 texts from all other clusters are randomly sampled.
Next, the answers to a question $q$ for these texts are predicted by LLMs.
After that, the pairs of the black-box embeddings from the sampled texts and their corresponding 1/0 answers are used as the training data to train a regression model.
Training details are provided in Appendix~\ref{apd:train_qa_models}.

The QA-based text embedding for a given input text is obtained by feeding a black-box embedding into the regression models for all questions.

\section{Experiments}
\label{sec:experiments}
This section demonstrates the effectiveness of our method by comparing it with the baseline.
We first describe the experimental settings, such as the baseline and hyperparameters~(\S~\ref{subsec:setup}).
After that, we explain the metrics for measuring the quality and interpretability of QA-based text embeddings~(\S~\ref{subsec:metrics}).
Finally, we evaluate the quality and interpretability of our method and the baseline on general-purpose tasks~(\S~\ref{subsec:result_and_discussion}).

\subsection{Experimental Setup}
\label{subsec:setup}
\begin{table}[!t]
    \centering
    \small
    \begin{tabular}{l|cccc|c}
        \toprule
        Level           & 1     & 2     & 3     & 4         & Baseline  \\
        \midrule
        \# Clusters     & ~~40  & 200   & 1,000 & ~~4,996   & ~~5,000   \\
        \# Questions    & 213   & 861   & 3,462 & 13,914    & 16,419    \\
        \bottomrule
    \end{tabular}
    \caption{Number of clusters and questions.}
    \label{tab:n_clusters_and_questions}
\end{table}

We use CQG-MBQA~\citep{sun2025a} as the baseline.
CQG-MBQA uses a single clustering step instead of the recursive clustering used in our method.
For the recursive clustering described in Section~\ref{subsec:q_gen}, we set the parameters $k_\text{first}$ and $k$ to 40 and 5, respectively.
We use \texttt{UAE-large-v1}\footnote{\url{https://huggingface.co/WhereIsAI/UAE-Large-V1}} for text embedding, \texttt{gpt-4o-mini}\footnote{\url{https://developers.openai.com/api/docs/models/gpt-4o-mini}\label{foot:model_gpt}} for question generation, and \texttt{mDeBERTa-base-v3}\footnote{\url{https://huggingface.co/MoritzLaurer/mDeBERTa-v3-base-xnli-multilingual-nli-2mil7}\label{foot:model_nli}} as the NLI model for question alignment.
Table~\ref{tab:n_clusters_and_questions} shows the number of clusters and questions generated by our method and the baseline.
The number of clusters at the fourth level is slightly smaller than 5,000 because some clusters at the third level contain fewer than five samples and therefore cannot be further divided into five clusters.
The number of generated questions increases in proportion to the number of clusters.
The distribution of the number of texts in each cluster is provided in Appendix~\ref{apd:analysis_clusters}.

Table~\ref{tab:hyperparameter} shows the hyperparameters for question generation in our method and the baseline.
For direct comparison, we use the same hyperparameter settings for the fourth level as for the baseline.
As the hierarchical level becomes shallower, we increase $n_p$ by 2 and $n_h$ and $n_e$ by 6 each.
This is because clusters at shallow levels contain more samples and therefore cover broader regions of the semantic space, but using too few contrastive samples for these clusters may cause the generated questions to focus on only a limited region of the semantic space.

Our method and the baseline are evaluated on the three tasks, STS~\citep{agirre-etal-2012-semeval}, text clustering~\citep{Aggarwal2012}, and information retrieval~\citep{thakur2021beir}, using the datasets included in the massive text embedding benchmark (MTEB)~\citep{muennighoff-etal-2023-mteb}.
Details of the datasets are provided in Appendix~\ref{apd:eval_data}.

\begin{table}[!t]
    \centering
    \small
    \begin{tabular}{ccccccccc}
        \toprule
        Level       & $n_p$ & $n_h$ & $n_e$ & $n_c$ & $p_p$ & $p_h$ & $p_e$ & $\theta$  \\
        \midrule
        1           & 12    & 36    & 36    & 3     & 5     & 3     & 2     & 0.8       \\
        2           & 10    & 30    & 30    & 3     & 5     & 3     & 2     & 0.8       \\
        3           & ~~8   & 24    & 24    & 3     & 5     & 3     & 2     & 0.8       \\
        4           & ~~6   & 18    & 18    & 3     & 5     & 3     & 2     & 0.8       \\
        \midrule
        Baseline    & ~~6   & 18    & 18    & 3     & 5     & 3     & 2     & 0.8       \\
        \bottomrule
    \end{tabular}
    \caption{Hyperparameters for question generation.}
    \label{tab:hyperparameter}
\end{table}

\subsection{Metrics}
\label{subsec:metrics}
\paragraph{Embedding Quality}
Based on the basic settings in MTEB~\citep{muennighoff-etal-2023-mteb}, the STS task uses Spearman’s rank correlation coefficient, the text clustering task uses the V-Measure, and the information retrieval task uses nDCG@10 as evaluation metrics for the quality of text embeddings.

\paragraph{Interpretability}
As a metric for measuring interpretability, we define cognitive load as the number of questions a human must ask during interpretation.
Specifically, it is defined for each of the three tasks as follows.

\noindent
\textbf{\underline{STS}}:~
Let $\mathcal{I}_{\text{sts}}$ be the indices of questions answered ``yes'' in both sentence vectors $\mathbf{v}_1$ and $\mathbf{v}_2$.
The cognitive load is defined as the size of $\mathcal{I}_{\text{sts}}$.
\begin{equation}
    \mathcal{I}_{\text{sts}} = \{ i \,|\, v_{1i} = 1 \, \land \, v_{2i} = 1\}
\end{equation}
\begin{equation}
    \mathrm{cl}_{\text{sts}}(\mathbf{v}_1, \mathbf{v}_2) = |\mathcal{I}_{\text{sts}}|
\end{equation}

\noindent
\textbf{\underline{Clustering}}:~
Let $\mathcal{I}_{\text{clustering}}$ be the indices of questions answered ``yes'' in all text embeddings in a cluster $\mathcal{V}$, then $\mathrm{cl}_{\text{clustering}}(\mathcal{V}) = |\mathcal{I}_{\text{clustering}}|$.
\begin{equation}
  \mathcal{I}_{\text{clustering}} = \{ i \,|\, \forall \mathbf{v}_j \in \mathcal{V}: v_{ji} = 1 \}
\end{equation}

\noindent
\textbf{\underline{Retrieval}}:~
Let $\mathcal{I}_{\text{retrieval}}$ be the indices of questions answered ``yes'' in both the query vector $\mathbf{u}$ and the top-1 passage $\mathbf{v}$, then $\mathrm{cl}_{\text{retrieval}}(\mathbf{u},\mathbf{v}) = |\mathcal{I}_{\text{retrieval}}|$.
\begin{equation}
  \mathcal{I}_{\text{retrieval}} = \{ i \,|\, u_i = 1 \, \land \, v_i = 1\}\\
\end{equation}

Here, the cognitive load for STS is the same as that defined by \citet{sun2025a}, while the cognitive loads for clustering and retrieval are newly added in this study.

Furthermore, we assume that question alignment reduces cognitive load by replacing the question with its aligned abstract question at the higher level, thereby reducing the number of questions a human must check.
We define the cognitive load of embeddings with question alignment as
\begin{equation}
  \mathrm{cl}^{\text{qa}}_{x}(\cdot) = |\{T_\text{question}[l][i] \,|\, i \ \in \mathcal{I}_x\}|,
\end{equation}
where $x \in \{ \text{sts, clustering, retrieval} \}$ and $l$ is the question level.

\begin{figure*}[!t]
    \centering
    \begin{minipage}[b]{0.25\linewidth}
        \centering
        \includegraphics[width=1.0\linewidth]{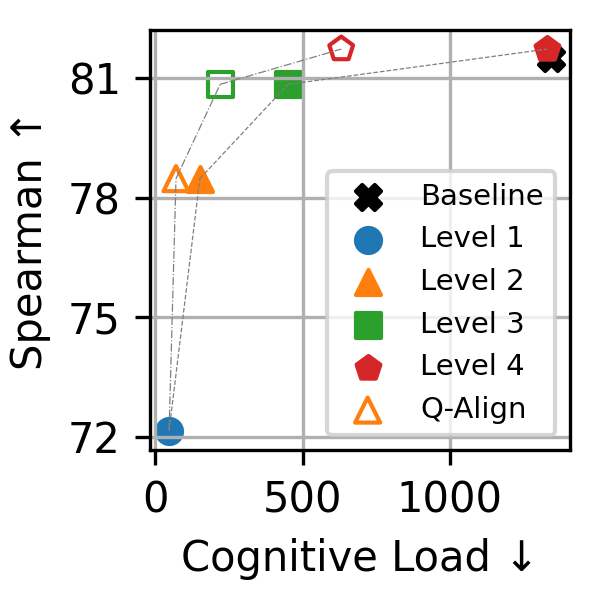}
        \subcaption{STS}
    \end{minipage}
    \hspace{1cm}
    \begin{minipage}[b]{0.25\linewidth}
        \centering
        \includegraphics[width=1.0\linewidth]{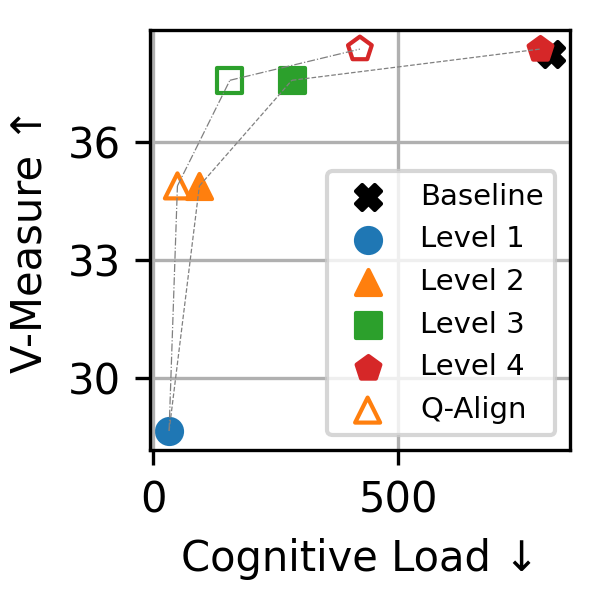}
        \subcaption{Clustering}
    \end{minipage}
    \hspace{1cm}
    \begin{minipage}[b]{0.25\linewidth}
        \centering
        \includegraphics[width=1.0\linewidth]{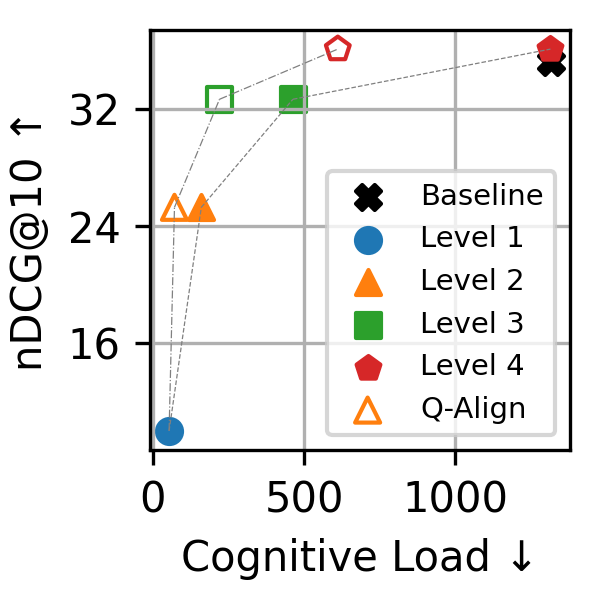}
        \subcaption{Retrieval}
    \end{minipage}
    \caption{Task performance vs. cognitive load ($\tau$ is 0.3). Q-Align indicates the results after question alignment, shown as white-filled shapes.}
    \label{fig:scores_and_cl_reduced}
\end{figure*}

\begin{figure*}[!t]
    \centering
    \begin{minipage}[b]{0.25\linewidth}
        \centering
        \includegraphics[width=1.0\linewidth]{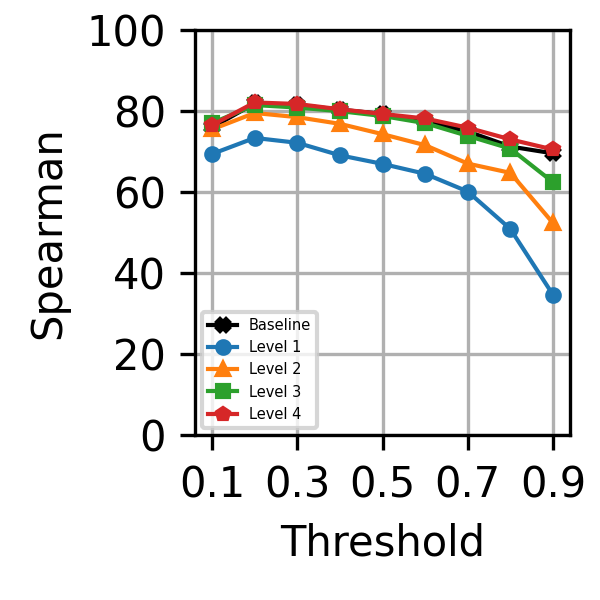}
        \subcaption{STS}
        \label{fig:thr_scores_sts}
    \end{minipage}
    \hspace{1cm}
    \begin{minipage}[b]{0.25\linewidth}
        \centering
        \includegraphics[width=1.0\linewidth]{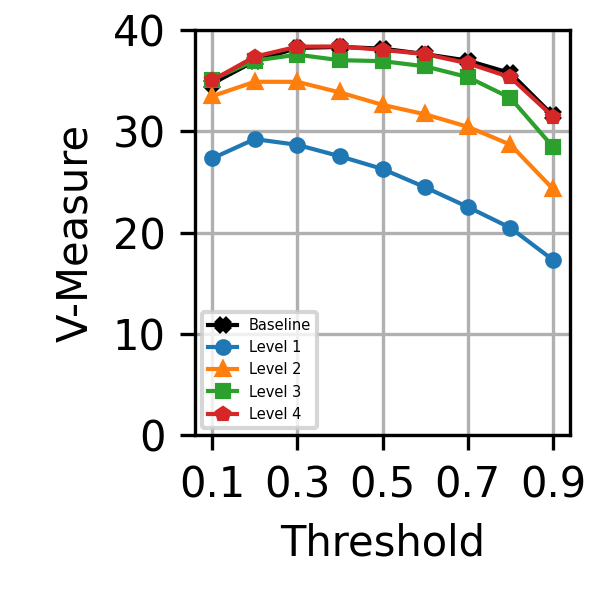}
        \subcaption{Clustering}
        \label{fig:thr_scores_clustering}
    \end{minipage}
    \hspace{1cm}
    \begin{minipage}[b]{0.25\linewidth}
        \centering
        \includegraphics[width=1.0\linewidth]{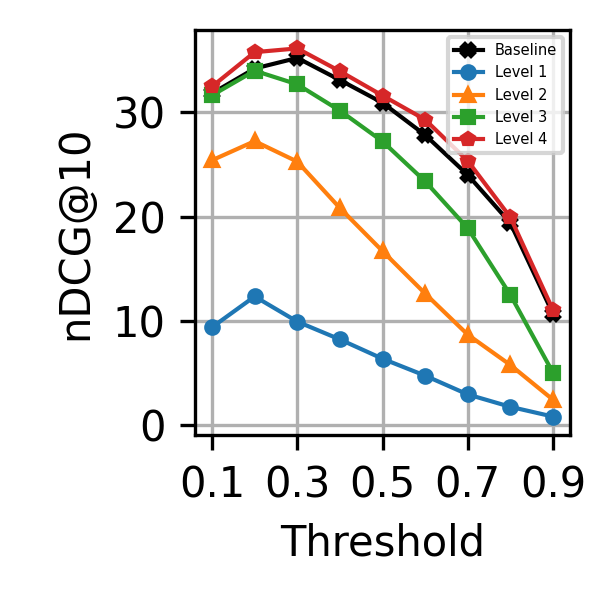}
        \subcaption{Retrieval}
        \label{fig:thr_scores_retrieval}
    \end{minipage}
    \caption{Task performance for the different threshold $\tau$.}
    \label{fig:thr_scores}
\end{figure*}

\begin{figure*}[!t]
    \centering
    \begin{minipage}[b]{0.25\linewidth}
        \centering
        \includegraphics[width=1.0\linewidth]{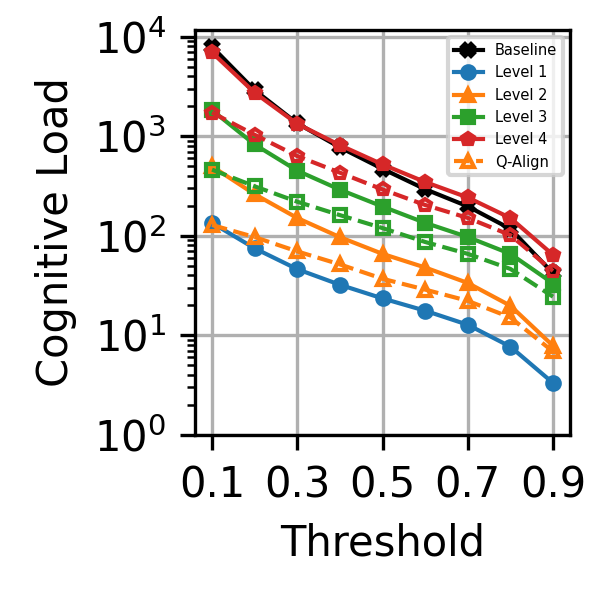}
        \subcaption{STS}
        \label{fig:thr_cl_reduced_sts}
    \end{minipage}
    \hspace{1cm}
    \begin{minipage}[b]{0.25\linewidth}
        \centering
        \includegraphics[width=1.0\linewidth]{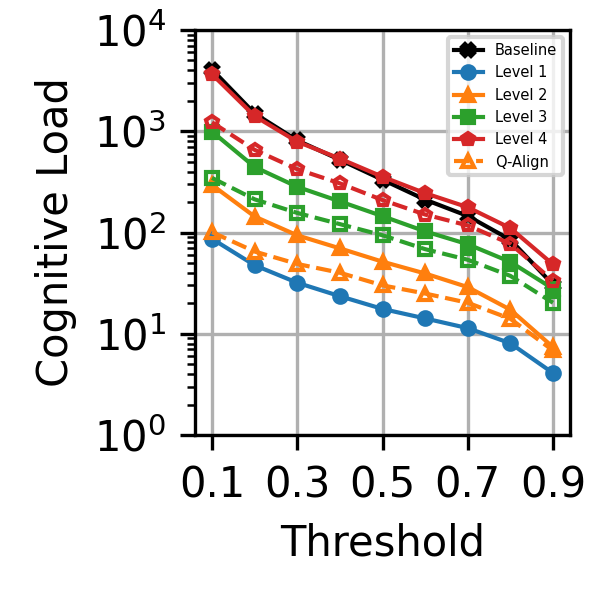}
        \subcaption{Clustering}
        \label{fig:thr_cl_reduced_clustering}
    \end{minipage}
    \hspace{1cm}
    \begin{minipage}[b]{0.25\linewidth}
        \centering
        \includegraphics[width=1.0\linewidth]{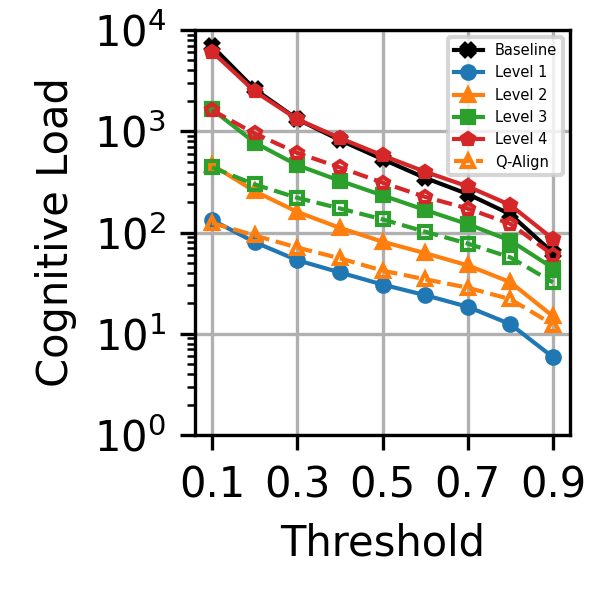}
        \subcaption{Retrieval}
        \label{fig:thr_cl_reduced_retrieval}
    \end{minipage}
    \caption{Cognitive load for the different threshold $\tau$. Q-Align indicates the results after question alignment, shown as white-filled shapes.}
    \label{fig:thr_cl_reduced}
\end{figure*}

\subsection{Result and Discussion}
\label{subsec:result_and_discussion}
\paragraph{Performance vs. Cognitive Load}
Figure~\ref{fig:scores_and_cl_reduced} shows the performance and cognitive load on the STS, text clustering, and information retrieval tasks when the threshold $\tau$ is 0.3.
Several common trends can be observed across the three tasks.
First, the performance and cognitive load of the fourth level are nearly identical to those of the baseline.
This suggests that questions generated through recursive clustering provide similar embedding quality and cognitive load to those generated using a single clustering step.
Second, there is a trade-off between task performance and cognitive load.
In other words, obtaining higher-quality text embeddings requires more questions, increasing cognitive load.
A user can choose the level of embeddings, considering a balance between the performance of the downstream task and the cognitive load of interpreting the model’s prediction.
In this experiment, since cognitive load increases significantly from the third level to the fourth level while performance improves only slightly, the third level appears to offer the best balance between task performance and cognitive load.
Finally, question alignment can reduce cognitive load while maintaining task performance.
As a result, the Pareto front of task performance and cognitive load is significantly improved, outperforming the baseline.
Results for other values of the threshold $\tau$ are given in Appendix~\ref{apd:full_results_performance_vs_cognitive_load}, indicating a similar trade-off between performance and cognitive load across all values.\footnote{The task performance shown in Figure~\ref{fig:scores_and_cl_reduced} and Appendix~\ref{apd:full_results_performance_vs_cognitive_load} is the average of the evaluation metrics across the datasets. The results for individual datasets are shown in Appendix~\ref{apd:performance_by_dataset}.}

\paragraph{Impact of Thresholds on Performance and Cognitive Load}
Figures~\ref{fig:thr_scores}~and~\ref{fig:thr_cl_reduced} show the task performance and cognitive load for the threshold $\tau$, respectively.
As shown in Figure~\ref{fig:thr_scores}, performance peaks around a threshold of 0.3 for all three tasks and gradually decreases thereafter.
Furthermore, the difference in performance between the third and fourth level is not particularly significant, indicating that high-quality text embeddings can be generated for general-purpose tasks even with the number and granularity of questions at the third level.
As shown in Figure~\ref{fig:thr_cl_reduced}, cognitive load decreases significantly as the threshold increases.
Furthermore, the effect of reduction of cognitive load by question alignment is greater when the level is deep or the threshold is low.
However, if the threshold is too low, the discrepancy between the model’s and human’s standards, such as a human interpreting a question the model answered ``yes'' to as ``no'', may become greater, potentially reducing the reliability of the generated text embeddings.

\section{Analysis}
This section analyzes our method from various perspectives: entailment relationships between aligned questions~(\S~\ref{subsec:entailment_questions}), the appropriate question granularity~(\S~\ref{subsec:granularity_questions}), and a case study~(\S~\ref{subsec:case_study}).

\subsection{Entailment Relationships between Aligned Questions}
\label{subsec:entailment_questions}
\begin{figure}[!t]
    \centering
    \includegraphics[width=0.75\linewidth]{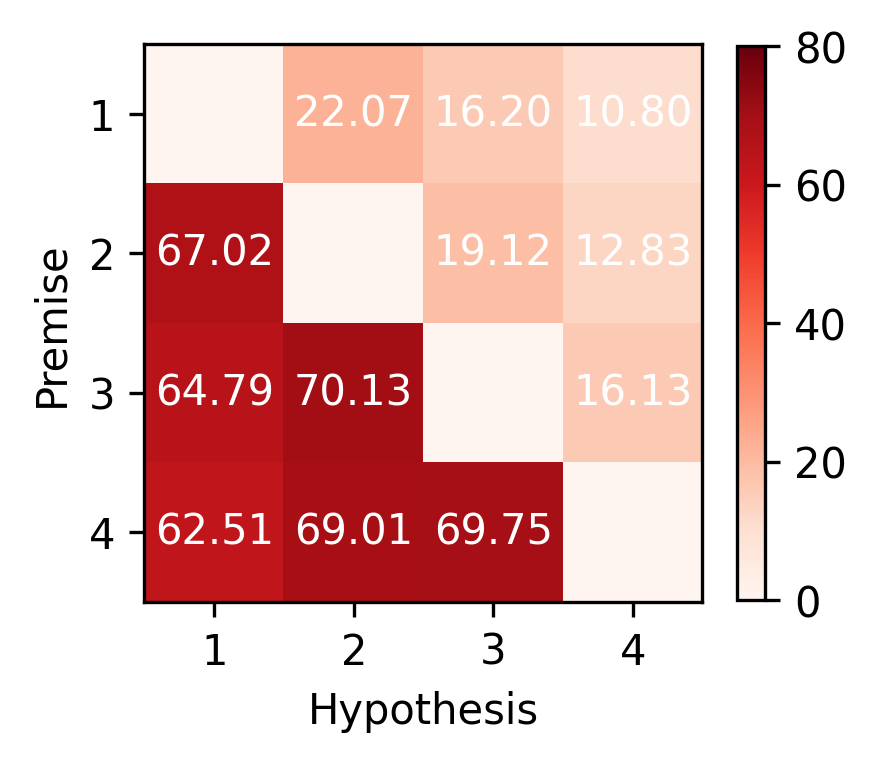}
    \caption{Proportion of cases where the probability of ``entailment'' exceeded 50\%.}
    \label{fig:entailment_proportion}
\end{figure}

To reduce cognitive load through question alignment, the target question needs to semantically summarize the source questions.
To verify this, the entailment relationship between the aligned questions is estimated.
Specifically, for the $i$-th source question $q^{l_s}_i$ at the level $l_s$ and the $j$-th target question $q^{l_t}_j$ at the level $l_t$ (where $l_s > l_t$), we convert them into declarative sentences $d^{l_s}_i$ and $d^{l_t}_j$.
After that, the probability $p(d^{l_t}_j|d^{l_s}_i)$ that $d^{l_s}_i$ entails $d^{l_t}_j$, and the probability $p(d^{l_s}_i|d^{l_t}_j)$ that $d^{l_t}_j$ entails $d^{l_s}_i$ is estimated by an NLI model\footref{foot:model_nli}.
If $p(d^{l_t}_j|d^{l_s}_i)$ is high (i.e., $q^{l_s}_i$ entails $q^{l_t}_j$) and $p(d^{l_s}_i|d^{l_t}_j)$ is largely lower than $p(d^{l_t}_j|d^{l_s}_i)$ (i.e., $q^{l_t}_j$ is more abstract than $q^{l_s}_i$), $q^{l_t}_j$ can be considered suitable for replacing $q^{l_s}_i$.
When the difference between $l_s$ and $l_t$ is 2 or greater, $j$ is determined by recursively consulting the alignment table $T_\text{question}$.
For example, if $l_s$ is 3 and $l_t$ is 1, $j$ is $T_\text{question}[2][T_\text{question}[3][i]]$.

Figure~\ref{fig:entailment_proportion} shows the proportion of combinations of questions across different levels for which the entailment probability exceeded 50\%.
For many level pairs, more than 67\% of lower-level questions entail upper-level questions.
On the other hand, the proportion of upper-level questions that entail lower-level questions is small.
These results indicate that the granularity of the questions becomes finer as the level deepens, while aligned questions preserve semantic coherence. Therefore, our method generates hierarchical questions in which upper-level questions summarize lower-level questions.
The distribution of the probability of entailment is provided in Appendix~\ref{apd:analysis_questions}.

\subsection{Impact of Granularity of Questions}
\label{subsec:granularity_questions}

\begin{figure}[!t]
    \begin{minipage}[b]{0.495\linewidth}
        \centering
        \includegraphics[width=1.0\linewidth]{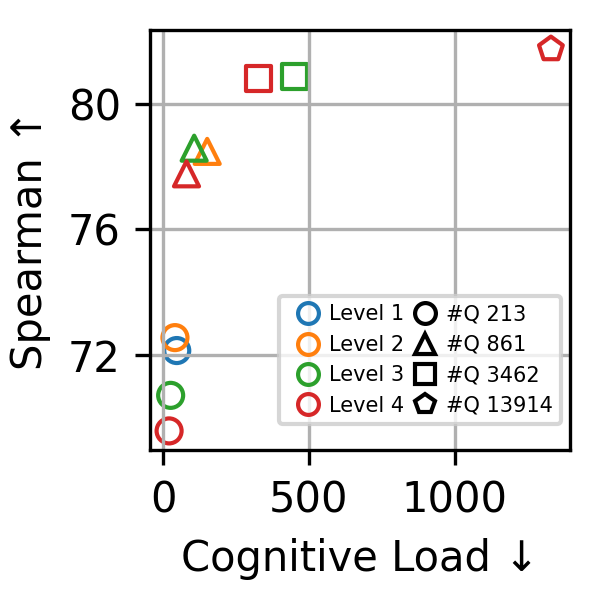}
        \subcaption{$\tau$ is 0.3}
        \label{fig:scores_and_cl_random_03}
    \end{minipage}
    \hfill
    \begin{minipage}[b]{0.495\linewidth}
        \centering
        \includegraphics[width=1.0\linewidth]{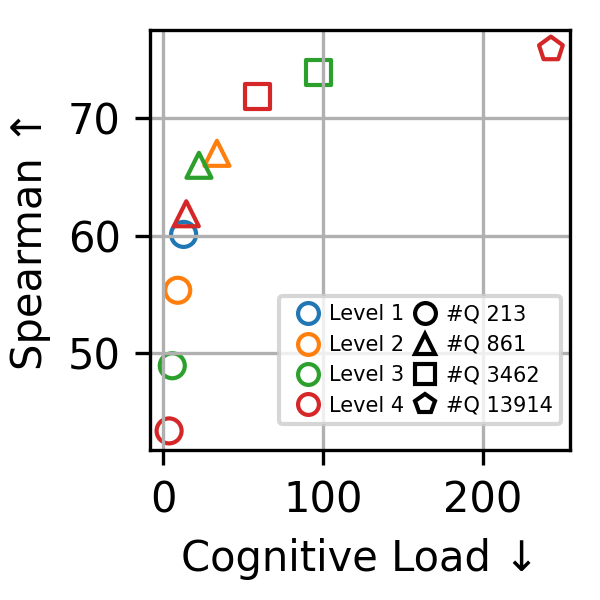}
        \subcaption{$\tau$ is 0.7}
        \label{fig:scores_and_cl_random_07}
    \end{minipage}
    \caption{Task performance vs. cognitive load when the number of questions is matched across levels by random sampling.}
    \label{fig:scores_and_cl_random}
\end{figure}

We investigate how question granularity affects task performance and cognitive load.
Figure~\ref{fig:scores_and_cl_reduced} shows that embeddings encoded by the questions at the fourth level achieve the highest performance, but this may stem from using the largest number of questions.
To evaluate the appropriateness of the questions themselves, questions from deeper levels are randomly chosen so that the number of sampled questions matches that of the upper levels, and the resulting text embeddings are evaluated using the same metrics as in Section~\ref{sec:experiments}.
Figure~\ref{fig:scores_and_cl_random} shows the mean Spearman's rank correlation coefficient and cognitive load for the STS task across five different seeds when the threshold $\tau$ is 0.3 and 0.7.
When $\tau$ is $0.3$, there is little difference in performance across question granularities.
In contrast, when $\tau$ is $0.7$, task performance at upper levels tends to be higher than that at lower levels with the same number of questions.
This trend is particularly pronounced when the number of questions is small.
When the threshold is high and ``yes'' answers are less likely, fine-grained questions are answered ``no'' in most cases, making the resulting text embeddings less informative, although cognitive cost is reduced.
In such cases, abstract questions are more appropriate for QA-based text embeddings.
Results for other values of $\tau$ and other tasks are provided in Appendix~\ref{apd:full_results_granularity_questions}.

\begin{table*}[!t]
    \centering
    \small
    \begin{tabular}{lccl}
        \toprule
        \multicolumn{4}{l}{\textbf{Text 1:} what level is blood pressure too low} \\
        \multicolumn{4}{p{15.5cm}}{\textbf{Text 2:} As long as you are not feeling dizzy or passing out, and have a decent quality of life, it is totally okay to have that level of blood pressure. There are a lot of normal people walking around with blood pressures of 90/60 on no medication and it is just normal for them. The only time doctors treat low blood pressure is when there are any symptoms, such as dizziness, light-headedness, or the experience of passing out. Any of these could result in a potentially disastrous fall, so if you experience one of these symptoms then you do want to tell your doctor right away.} \\
        \midrule
        \textbf{Source Question} & \textbf{A1} & \textbf{A2} & \textbf{Target Question} \\
        \rowcolor{gray!10} Is the article likely to be used in academic settings or coursework? & Yes & Yes & Is the article concerned with \\
        \rowcolor{gray!10} Is there a clear intent to explain a concept or process? & Yes & Yes & enhancing understanding or skills? \\
        Is the information presented in a clinical or educational context? & Yes & Yes & Does the article discuss physical or \\
        Is the information provided related to health or medical conditions? & Yes & Yes & anatomical concepts? \\
        \rowcolor{gray!10} Is there mention of commonly accepted blood pressure thresholds? & No & Yes & \\
        Is there an absence of alarming or unsafe conditions in the description? & Yes & No & \\
        \bottomrule
    \end{tabular}
    \caption{Case study of interpreting text embeddings using question alignment. A1 and A2 denote the answers of the model for Text 1 and Text 2, respectively.}
    \label{tab:case_study}
\end{table*}

\subsection{Case Study}
\label{subsec:case_study}
As a case study, we present an example of interpreting text embeddings using our method.
The two input texts are sampled from the MS MARCO dataset~\citep{bajaj2018msmarcohumangenerated} for the retrieval task, and the threshold $\tau$ is set to 0.5.
Questions at the fourth level for which the model answers ``yes''  for both texts are replaced with their aligned questions at the third level.
The results shown in Table~\ref{tab:case_study} illustrate the interpretability of our method.
For questions related to concepts shared by two texts (e.g., education and medical care), the models answer ``yes'' for both.
In contrast, for questions related to concepts that differ between the two texts (e.g., specific numerical thresholds and unsafe conditions), the models answer ``yes'' only for the corresponding text.
Furthermore, the aligned questions summarize the questions for which the model answers ``yes'' for both texts.
This makes it easier for humans to interpret the text embeddings.

\section{Related Work}
\paragraph{General Text Embeddings}
Text embeddings have been actively studied in the field of NLP~\citep{conneau-etal-2017-supervised, reimers-gurevych-2019-sentence, jiang-etal-2022-promptbert}.
Many recent text embedding models use Transformer encoders~\citep{devlin-etal-2019-bert, liu2019robertarobustlyoptimizedbert, vaswani2023attentionneed} trained using a contrastive learning framework~\citep{pmlr-v119-chen20j, gao-etal-2021-simcse, wang2024textembeddingsweaklysupervisedcontrastive}.
More recently, methods that use the hidden states of LLMs as text embeddings have also been proposed.
These include methods that make the attention in LLMs bidirectional and apply contrastive learning~\citep{behnamghader2024llmvec, muennighoff2025generative, lee2025nvembed}, as well as methods that insert input text to specific prompt templates without updating the parameters of the LLMs~\citep{jiang-etal-2024-scaling, lei-etal-2024-meta, lin2026reverseeolimprovingtrainingfreetext}.

\paragraph{Interpretable Text Embeddings}
Various methods have been proposed for interpretable text embeddings~\citep{hewitt-etal-2023-backpack, tehenan2025mechanisticdecompositionsentencerepresentations, anand-etal-2026-ibert}.
Many of these methods assign specific roles to each dimension of the text embeddings~\citep{mcinerney-etal-2023-chill, patel-etal-2023-learning, shin2026aligning}.
\citet{opitz-frank-2022-sbert} mapped each dimension of the text embeddings to an abstract meaning representation metrics.
\citet{wang-etal-2025-ldir} mapped each dimension of the text embeddings to an anchor text and defined its value as the similarity between the corresponding anchor text and the input sentence.
However, because these methods produce dense vectors, their interpretations may vary across individuals.
In contrast, QA-based text embeddings~\citep{benara2024crafting, sun2025a, tang2026askingrightquestionsontologygrounded} are sparse, making such ambiguity less likely.

\section{Conclusion}
In this paper, we proposed an interpretable text embedding framework with hierarchical questions to improve the interpretability of QA-based text embeddings.
Experiments across multiple tasks showed that our method improved the Pareto front derived from the trade-off between task performance and cognitive load.
Our analysis further showed that the aligned questions frequently exhibited entailment relationships, and the questions in shallower levels can be more effective when only a limited number of questions are available.


\section*{Limitations}
Although our method reduces cognitive load during interpretation, manually inspecting all generated questions remains difficult due to their large number.
Therefore, our qualitative analysis is limited to sampled questions, and a more comprehensive human evaluation is needed to assess the quality and interpretability of the generated questions.

Another limitation is that our method focuses on improving the Pareto front derived from the trade-off between task performance and cognitive load rather than directly improving task performance.
Future work could study how the question hierarchy can be used to improve embedding quality while keeping the embeddings interpretable.

In addition, we do not evaluate our method on multiple languages.
The quality and granularity of generated questions may differ across languages, so it is still unclear whether our method works well in other languages.
Evaluating our method in multilingual settings is therefore an important direction for future work.


\bibliography{custom,anthology-1,anthology-2}


\appendix

\section{Prompts}
\label{apd:prompts}
The prompt templates used in our experiments are shown in Figures~\ref{fig:qa-prompt}--\ref{fig:contrastive-question-prompt}.
We use \texttt{gpt-4o-mini}\footref{foot:model_gpt} for all LLM-based processing described in this section.

\subsection{Prompt: QA-based Text Embedding}
Figure~\ref{fig:qa-prompt} shows the prompt template used to obtain QA-based text embeddings.
The prompt takes two inputs: an input text and a set of yes/no questions.
The LLM is instructed to answer each question with either ``yes'' or ``no'' in the same order as the questions.
The resulting answers are used as labels for the corresponding text--question pairs.

\begin{figure}[!ht]
\centering

\begin{tcolorbox}[
    colback=white,
    colframe=black!60,
    boxrule=0.5pt,
    arc=2mm,
    left=2mm,
    right=2mm,
    top=2mm,
    bottom=2mm,
    width=\linewidth,
    fontupper=\small\ttfamily,
]
\raggedright
\setlength{\parindent}{0pt}

Evaluate the following text chunk based on the yes/no questions provided.

\medskip

**Text Chunk:**

\{text chunk\}

\medskip

**Questions:**

\{questions\}

\medskip

**Instruction for the model:** Please read the provided text chunk and answer each of the questions with either ``yes'' or ``no''. Format the responses as follows:

1. yes/no\\
2. yes/no

\end{tcolorbox}

\caption{Prompt template for QA-based text embedding.}
\label{fig:qa-prompt}
\end{figure}

\subsection{Prompt: Question-to-Declarative Conversion}
Figure~\ref{fig:declarative-prompt} shows the prompt template used for question-to-declarative conversion.
The prompt takes a yes/no question as input.
The LLM is instructed to convert the question into a corresponding declarative sentence and to output only the converted sentence.

\begin{figure}[!ht]
\centering

\begin{tcolorbox}[
    colback=white,
    colframe=black!60,
    boxrule=0.5pt,
    arc=2mm,
    left=2mm,
    right=2mm,
    top=2mm,
    bottom=2mm,
    width=\linewidth,
    fontupper=\small\ttfamily,
]
\raggedright
\setlength{\parindent}{0pt}

Convert the following question into a declarative sentence.

\medskip

**Question:** 

\{question\}

\medskip

**Instruction:** Please output only the declarative sentence derived from the given interrogative sentence.

\end{tcolorbox}

\caption{Prompt template for question-to-declarative conversion.}
\label{fig:declarative-prompt}
\end{figure}

\subsection{Prompt: Contrastive Question Generation}

Figure~\ref{fig:contrastive-question-prompt} shows the prompt template used for contrastive question generation.
The prompt takes two sets of articles as input: positive articles and negative articles.
The LLM is instructed to generate 10 concise yes/no questions that yield ``yes'' for all positive articles and ``no'' for all negative articles.
To encourage simple question structures, we additionally instruct the LLM to avoid using ``and'' or ``or'' unless necessary.

\begin{figure}[!ht]
\centering

\begin{tcolorbox}[
    colback=white,
    colframe=black!60,
    boxrule=0.5pt,
    arc=2mm,
    left=2mm,
    right=2mm,
    top=2mm,
    bottom=2mm,
    width=\linewidth,
    fontupper=\small\ttfamily,
]
\raggedright
\setlength{\parindent}{0pt}

Generate 10 simple yet insightful yes/no questions that determine properties of an article, where for all questions, the answer will be ``yes'' for ALL the positive articles and ``no'' for ALL the negative articles. Keep questions concise and avoid using complex sentence structures with ``and'' or ``or'' unless necessary.

\medskip

**Positive Articles:** 

\{positive articles\}

\medskip

**Negative Articles:** 

\{negative articles\}

\medskip

**Instruction:** Based on the excerpts provided, generate 10 simple yet insightful yes/no questions that can accurately differentiate the positive articles from the negative articles. Each question should be concise and framed in such a way that it will elicit a ``yes'' response for ALL positive articles and a ``no'' response for ALL negative articles. Avoid using complex sentence structures with ``and'' or ``or'' unless absolutely necessary. Format the questions in a numbered list as shown below:

1. [First simple yes/no question]\\
2. [Second simple yes/no question]

\end{tcolorbox}

\caption{Prompt template for contrastive question generation.}
\label{fig:contrastive-question-prompt}
\end{figure}

\section{Details of Filtering}
\label{apd:filtering}
Algorithm~\ref{alg:q_filtering} shows the detailed procedure for question filtering.
It consists of two steps, and $\mathbf{R}^l_\text{step1}$ and $\mathbf{R}^l_\text{step2}$ are collections of sets of questions after filtering in Step 1 and Step 2, respectively.
It is basically the same as the method proposed by \citet{sun2025a}.
However, in this study, the flag $top\_t$ at line 14 is set to false to avoid selecting the top $t$ questions, because it may eliminate promising candidates.

{
\algrenewcommand{\algorithmicindent}{1.0em}
\begin{algorithm}[!ht]
\caption{Question Filtering}
\label{alg:q_filtering}
\begin{algorithmic}[1]
\State \textbf{\# Step1. Filtering for each cluster}
\State $\mathbf{R}^l_\text{step1} \gets \emptyset$
\For{$\mathcal{R} \in \mathbf{R}^l$}
    \State $\mathcal{R}_\text{exclude} \gets \emptyset$
    \State $\mathcal{A} \gets \{(q_i, q_j) \in \mathcal{R} \times \mathcal{R} \;|\; q_i \neq q_j\}$
    \For{$q_i, q_j \in \mathcal{A}$}
        \If{$\mathrm{sim}(q_i, q_j) > \theta$}
            \If{$\mathrm{quality}(q_i) < \mathrm{quality}(q_j)$}
                \State $\mathcal{R}_\text{exclude} \gets \mathcal{R}_\text{exclude} \cup \{q_i\}$
            \EndIf
        \EndIf
    \EndFor
    \State $\mathcal{R} \gets \mathcal{R} \setminus \mathcal{R}_\text{exclude}$
    \If{$top\_t$}
        \State $\mathcal{R} \gets \mathrm{top\_t}(\mathcal{R}, t)$
    \EndIf
    \State $\mathbf{R}^l_\text{step1} \gets \mathbf{R}^l_\text{step1} \cup \{\mathcal{R}\}$
\EndFor
\\
\State \textbf{\# Step2. Filtering across clusters}
\State $\mathbf{R}^l_\text{step2} \gets \emptyset$
\For{$\mathcal{R}_\text{tar} \in \mathbf{R}^l_\text{step1}$}
    \State $\mathcal{R}_\text{exclude} \gets \emptyset$
    \For{$\mathcal{R}_\text{ref} \in \mathbf{R}^l_\text{step1}$}
        \State $\mathcal{A} \gets \{(q_i, q_j) \in \mathcal{R}_\text{tar} \times \mathcal{R}_\text{ref} \;|\; q_i \neq q_j\}$
        \For{$q_i, q_j \in \mathcal{A}$}
            \If{$\mathrm{sim}(q_i, q_j) > \theta$}
                \If{$\mathrm{quality}(q_i) < \mathrm{quality}(q_j)$}
                    \State $\mathcal{R}_\text{exclude} \gets \mathcal{R}_\text{exclude} \cup \{q_i\}$
                \EndIf
            \EndIf
        \EndFor
    \EndFor
    \State $\mathcal{R}_\text{tar} \gets \mathcal{R}_\text{tar} \setminus \mathcal{R}_\text{exclude}$
    \State $\mathbf{R}^l_\text{step2} \gets \mathbf{R}^l_\text{step2} \cup \{\mathcal{R}_\text{tar}\}$
\EndFor
\State \Return $\mathbf{R}^l_\text{step2}$
\end{algorithmic}
\end{algorithm}
}

\section{Details of Training QA Models}
\label{apd:train_qa_models}
The structure of the QA model is the same as that of \citet{sun2025a}.
The regression model is a multilayer perceptron that passes a 1024-dimensional vector through a (1024, 8) linear layer, a ReLU activation function, and an (8, 1) linear layer, in that order.
The output logit is passed through a sigmoid function to obtain a probability of a ``yes'' answer.
For training hyperparameters, we set the batch size to 20, the number of epochs to 10, and the learning rate to 1e-3.
The model is trained using NVIDIA DGX Spark (128GB CPU).

\section{Analysis of Clusters}
\label{apd:analysis_clusters}
\begin{figure}[!ht]
    \centering
    \includegraphics[width=0.5\linewidth]{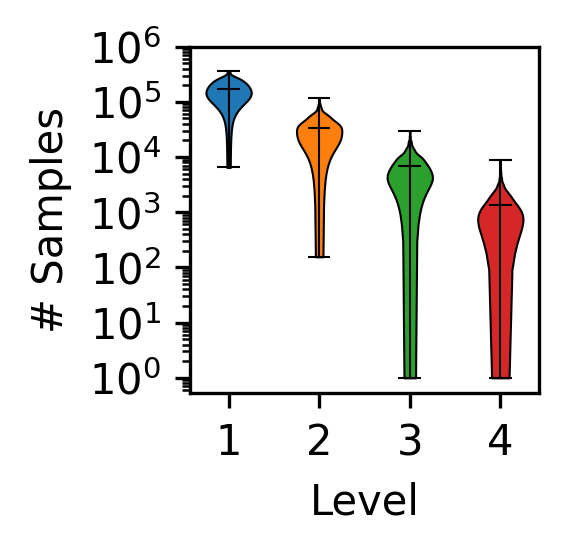}
    \caption{Distribution of the number of texts in each cluster.}
    \label{fig:n_samples}
\end{figure}

The distribution of the number of texts in each cluster is shown in Figure~\ref{fig:n_samples}.

\section{Evaluation Data}
\label{apd:eval_data}
The evaluation data for each task is as follows:
\begin{description}
    \item[STS]
    Seven datasets: SemEval STS tasks 2012-2016~\citep{agirre-etal-2012-semeval, agirre-etal-2013-sem, agirre-etal-2014-semeval, agirre-etal-2015-semeval, agirre-etal-2016-semeval}, STS Benchmark~\citep{cer-etal-2017-semeval}, and SICK-Relatedness~\citep{marelli-etal-2014-sick}.
    \item[Clustering]
    Seven datasets: TwentyNewsgroupsClustering\footnote{\url{https://huggingface.co/datasets/mteb/twentynewsgroups-clustering}}, StackExchangeClusteringP2P\footnote{\url{https://huggingface.co/datasets/mteb/stackexchange-clustering-p2p}}, BiorxivClusteringP2P\footnote{\url{https://huggingface.co/datasets/mteb/biorxiv-clustering-p2p}}, BiorxivClusteringS2S\footnote{\url{https://huggingface.co/datasets/mteb/biorxiv-clustering-s2s}}, MedrxivClusteringP2P\footnote{\url{https://huggingface.co/datasets/mteb/medrxiv-clustering-p2p}}, MedrxivClusteringS2S\footnote{\url{https://huggingface.co/datasets/mteb/medrxiv-clustering-s2s}}, and RedditClusteringP2P\footnote{\url{https://huggingface.co/datasets/mteb/reddit-clustering-p2p}}.
    \item[Retrieval]
    Six datasets: MS MARCO~\citep{bajaj2018msmarcohumangenerated}, ArguAna~\citep{wachsmuth-etal-2018-retrieval}, FiQA-2018~\citep{10.1145/3184558.3192301}, NFCorpus~\citep{boteva2016}, SCIDOCS~\citep{cohan-etal-2020-specter}, and SciFact~\citep{wadden-etal-2020-fact}.
\end{description}

\section{Analysis of Questions}
\label{apd:analysis_questions}

\subsection{Length of Questions}
\begin{figure}[!ht]
    \centering
    \includegraphics[width=0.5\linewidth]{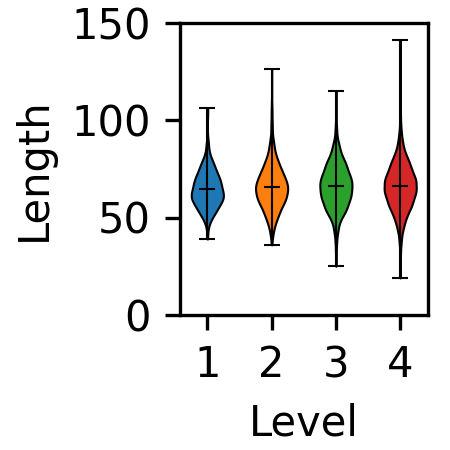}
    \caption{Distribution of the number of characters in questions.}
    \label{fig:length}
\end{figure}

The distribution of the number of characters in questions is shown in Figure~\ref{fig:length}.
Even across different levels, the length of questions does not vary much.

\subsection{Concreteness of Questions}
\begin{figure}[!ht]
    \centering
    \includegraphics[width=0.5\linewidth]{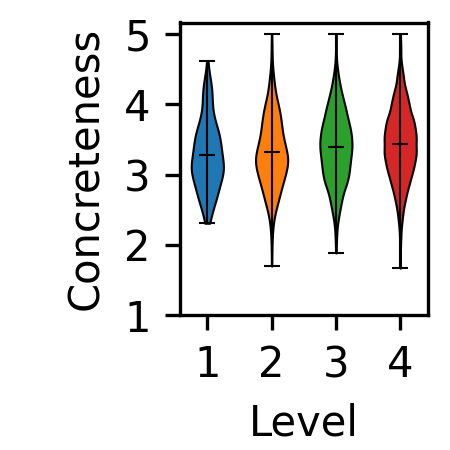}
    \caption{Distribution of the concreteness of questions.}
    \label{fig:concreteness}
\end{figure}

We calculate the concreteness of questions using wordtangible\footnote{\url{https://github.com/jrrobison1/wordtangible}}.
In wordtangible, a predefined dictionary, where the keys are words and the values are their concreteness scores, is used to compute the average concreteness of words as the concreteness of the sentence.
In the original wordtangible setting, words not included in the dictionary are ignored.
However, we believe that words not in the dictionary, such as proper nouns, should be considered concrete, so we assign them the maximum concreteness score (5 points).
The distribution of the concreteness of questions is shown in Figure~\ref{fig:concreteness}.
As the level becomes deeper, the average concreteness tends to increase.

\subsection{Probability of Entailment}
\begin{figure}[!ht]
    \centering
    \includegraphics[width=1.0\linewidth]{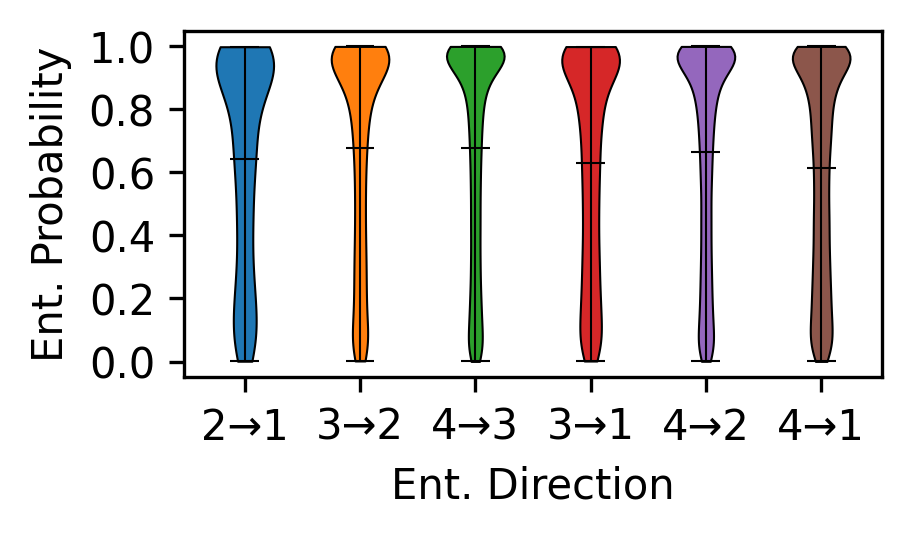}
    \caption{Distribution of probabilities of entailment (deeper-level $\rightarrow$ shallower-level).}
    \label{fig:prob1}
\end{figure}

\begin{figure}[!ht]
    \centering
    \includegraphics[width=1.0\linewidth]{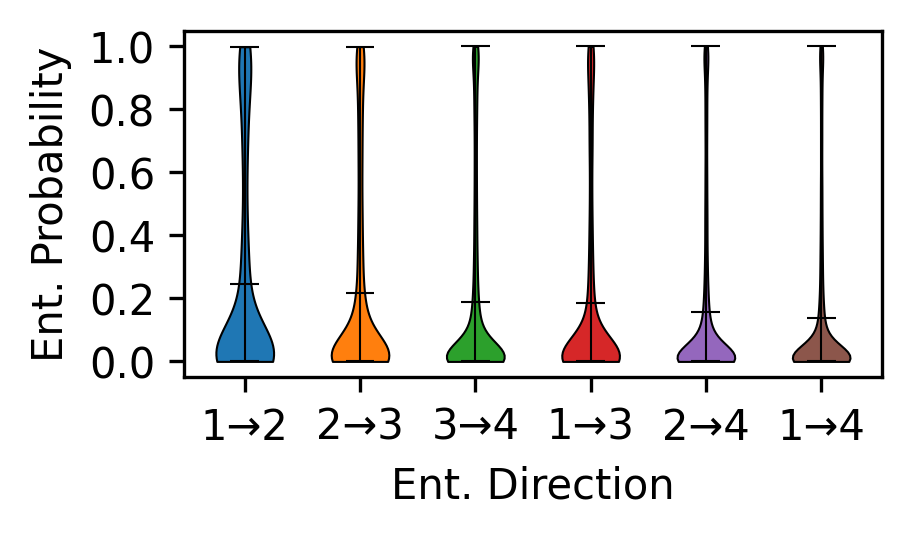}
    \caption{Distribution of probabilities of entailment (shallower-level $\rightarrow$ deeper-level).}
    \label{fig:prob2}
\end{figure}

Figure~\ref{fig:prob1} shows the distribution of entailment probabilities when the deeper-level sentences are the premise and the shallower-level sentences are the hypothesis, while Figure~\ref{fig:prob2} shows the distribution of entailment probabilities when the premise and hypothesis are swapped.

\section{Full Results of Performance vs. Cognitive Load}
\label{apd:full_results_performance_vs_cognitive_load}
\begin{figure*}[!ht]
    \centering
    \begin{minipage}[b]{0.25\linewidth}
        \centering
        \includegraphics[width=1.0\linewidth]{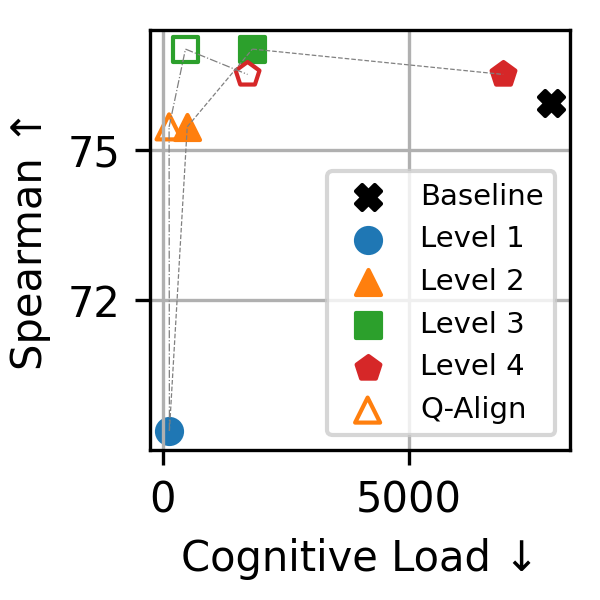}
        \subcaption{$\tau$ is 0.1}
    \end{minipage}
    \hspace{1cm}
    \begin{minipage}[b]{0.25\linewidth}
        \centering
        \includegraphics[width=1.0\linewidth]{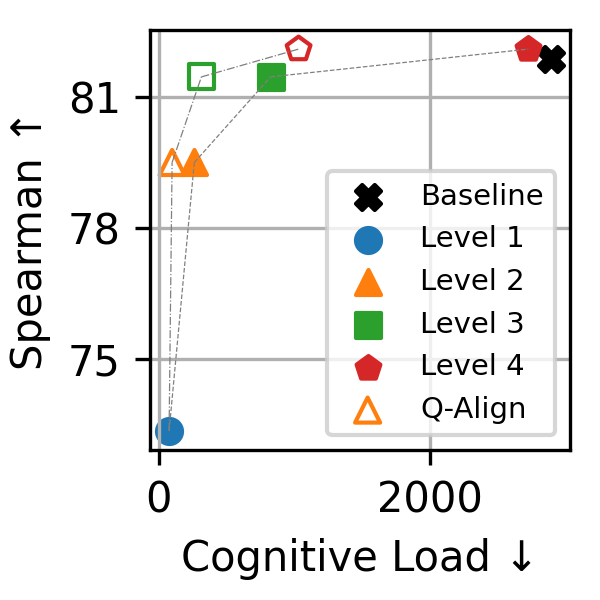}
        \subcaption{$\tau$ is 0.2}
    \end{minipage}
    \hspace{1cm}
    \begin{minipage}[b]{0.25\linewidth}
        \centering
        \includegraphics[width=1.0\linewidth]{figures/scores_and_cl_reduced/+2/with_t=False/sts/3.png}
        \subcaption{$\tau$ is 0.3}
    \end{minipage}
    \\\vspace{1cm}
    \begin{minipage}[b]{0.25\linewidth}
        \centering
        \includegraphics[width=1.0\linewidth]{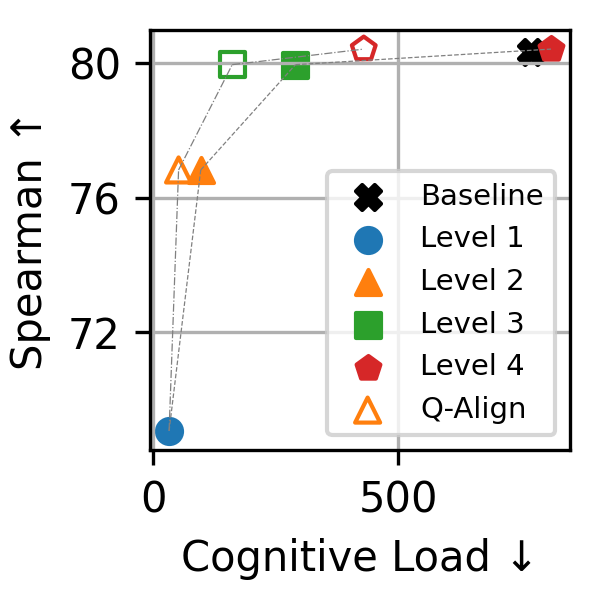}
        \subcaption{$\tau$ is 0.4}
    \end{minipage}
    \hspace{1cm}
    \begin{minipage}[b]{0.25\linewidth}
        \centering
        \includegraphics[width=1.0\linewidth]{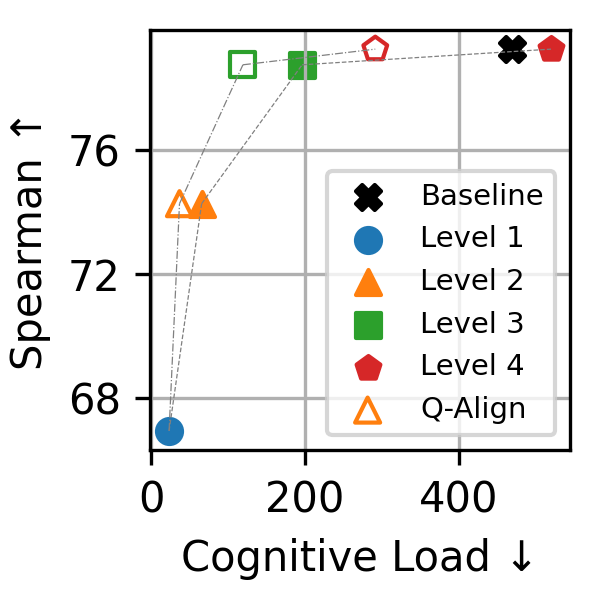}
        \subcaption{$\tau$ is 0.5}
    \end{minipage}
    \hspace{1cm}
    \begin{minipage}[b]{0.25\linewidth}
        \centering
        \includegraphics[width=1.0\linewidth]{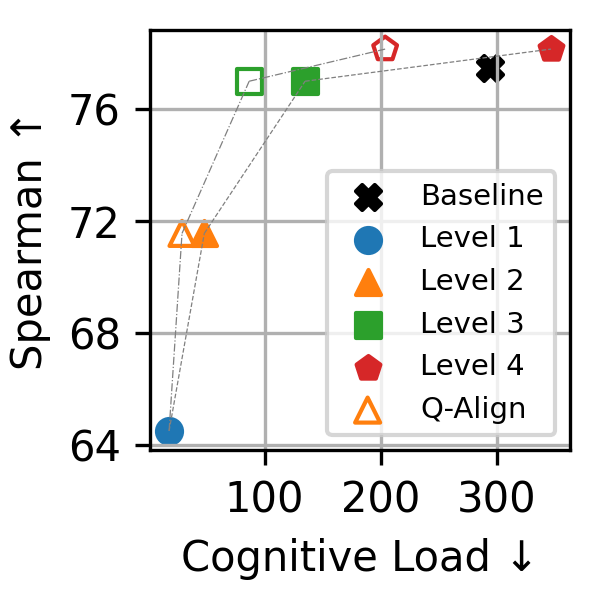}
        \subcaption{$\tau$ is 0.6}
    \end{minipage}
    \\\vspace{1cm}
    \begin{minipage}[b]{0.25\linewidth}
        \centering
        \includegraphics[width=1.0\linewidth]{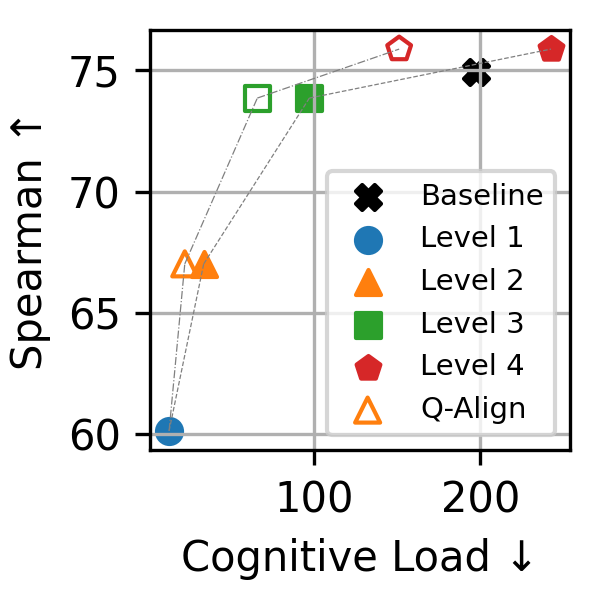}
        \subcaption{$\tau$ is 0.7}
    \end{minipage}
    \hspace{1cm}
    \begin{minipage}[b]{0.25\linewidth}
        \centering
        \includegraphics[width=1.0\linewidth]{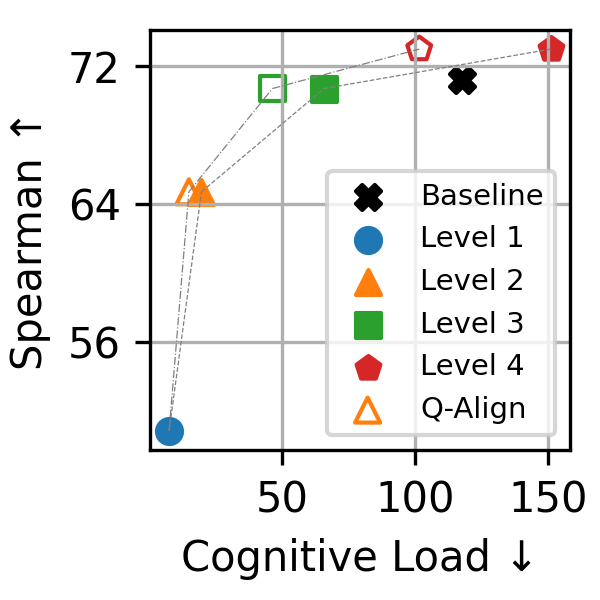}
        \subcaption{$\tau$ is 0.8}
    \end{minipage}
    \hspace{1cm}
    \begin{minipage}[b]{0.25\linewidth}
        \centering
        \includegraphics[width=1.0\linewidth]{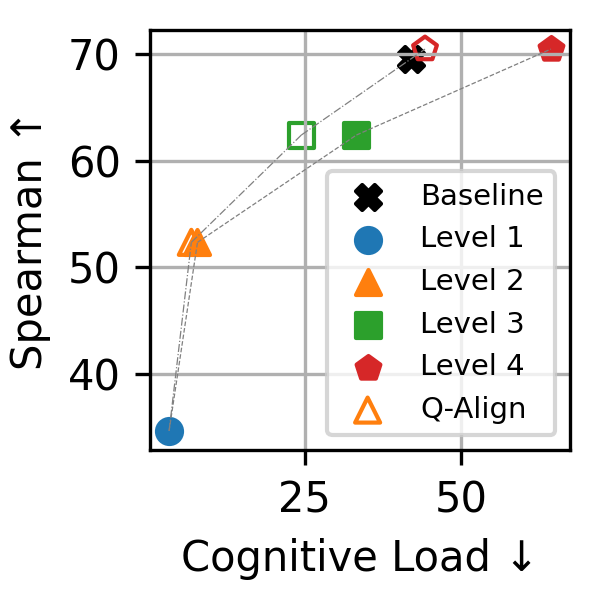}
        \subcaption{$\tau$ is 0.9}
    \end{minipage}
    \caption{Task performance vs. cognitive load on the STS task.}
    \label{fig:scores_and_cl_reduced_sts}
\end{figure*}

\begin{figure*}[!ht]
    \centering
    \begin{minipage}[b]{0.25\linewidth}
        \centering
        \includegraphics[width=1.0\linewidth]{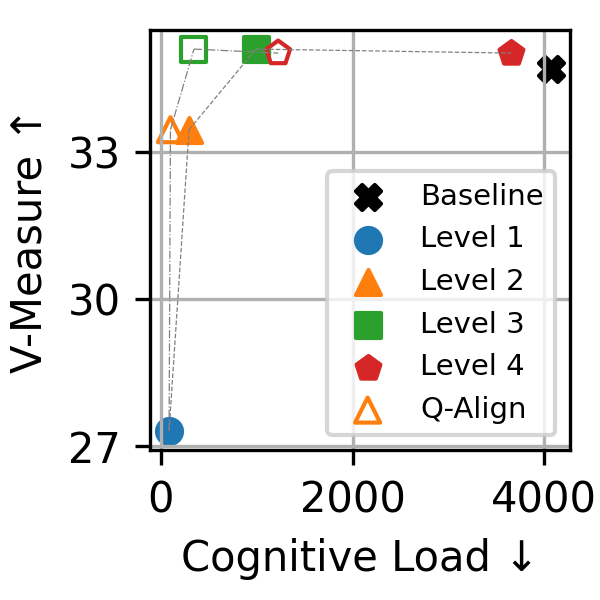}
        \subcaption{$\tau$ is 0.1}
    \end{minipage}
    \hspace{1cm}
    \begin{minipage}[b]{0.25\linewidth}
        \centering
        \includegraphics[width=1.0\linewidth]{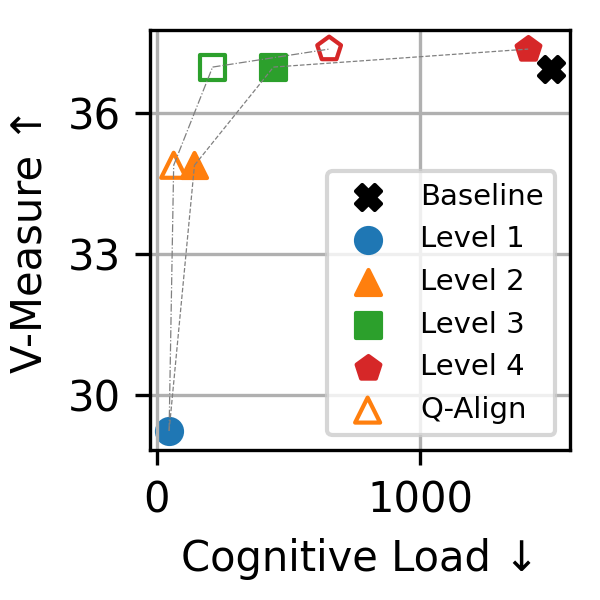}
        \subcaption{$\tau$ is 0.2}
    \end{minipage}
    \hspace{1cm}
    \begin{minipage}[b]{0.25\linewidth}
        \centering
        \includegraphics[width=1.0\linewidth]{figures/scores_and_cl_reduced/+2/with_t=False/clustering/3.png}
        \subcaption{$\tau$ is 0.3}
    \end{minipage}
    \\\vspace{1cm}
    \begin{minipage}[b]{0.25\linewidth}
        \centering
        \includegraphics[width=1.0\linewidth]{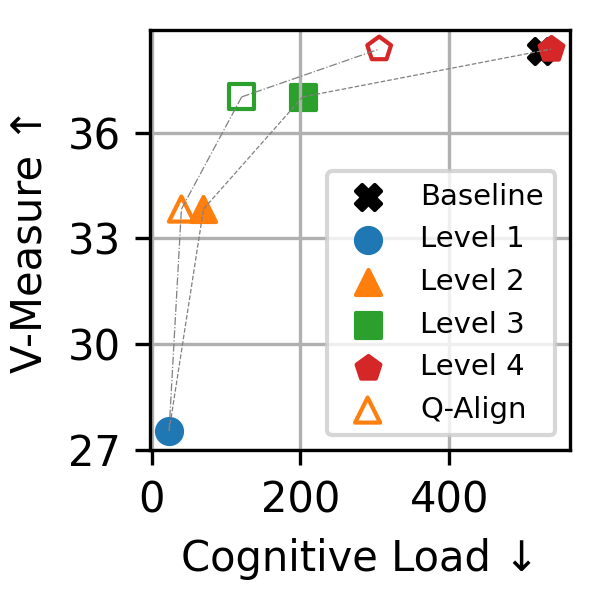}
        \subcaption{$\tau$ is 0.4}
    \end{minipage}
    \hspace{1cm}
    \begin{minipage}[b]{0.25\linewidth}
        \centering
        \includegraphics[width=1.0\linewidth]{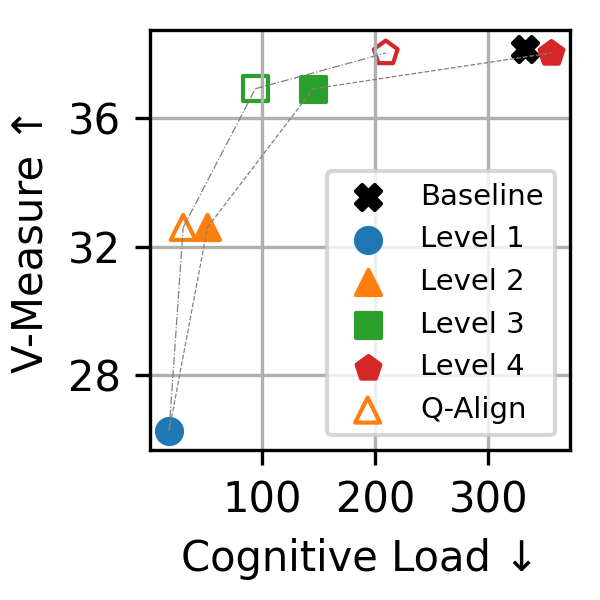}
        \subcaption{$\tau$ is 0.5}
    \end{minipage}
    \hspace{1cm}
    \begin{minipage}[b]{0.25\linewidth}
        \centering
        \includegraphics[width=1.0\linewidth]{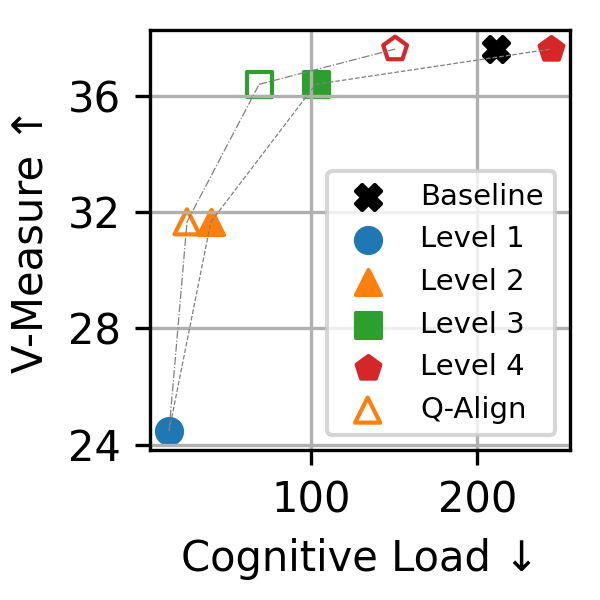}
        \subcaption{$\tau$ is 0.6}
    \end{minipage}
    \\\vspace{1cm}
    \begin{minipage}[b]{0.25\linewidth}
        \centering
        \includegraphics[width=1.0\linewidth]{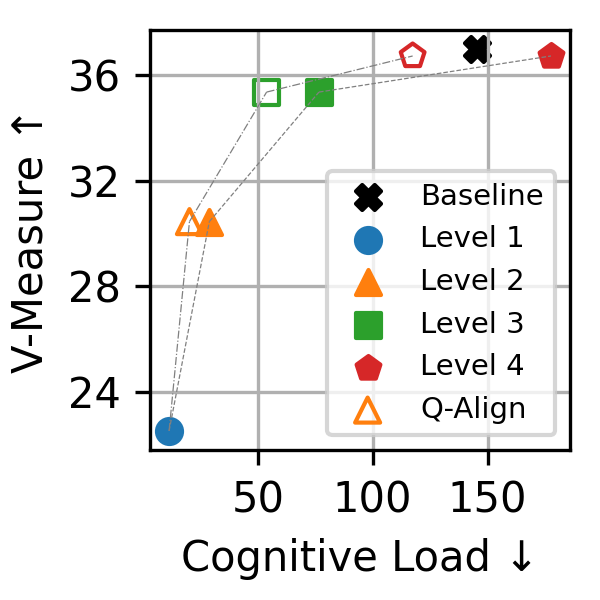}
        \subcaption{$\tau$ is 0.7}
    \end{minipage}
    \hspace{1cm}
    \begin{minipage}[b]{0.25\linewidth}
        \centering
        \includegraphics[width=1.0\linewidth]{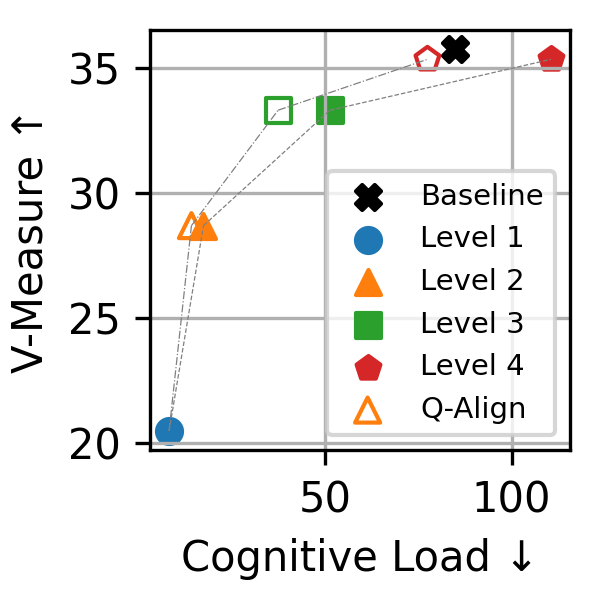}
        \subcaption{$\tau$ is 0.8}
    \end{minipage}
    \hspace{1cm}
    \begin{minipage}[b]{0.25\linewidth}
        \centering
        \includegraphics[width=1.0\linewidth]{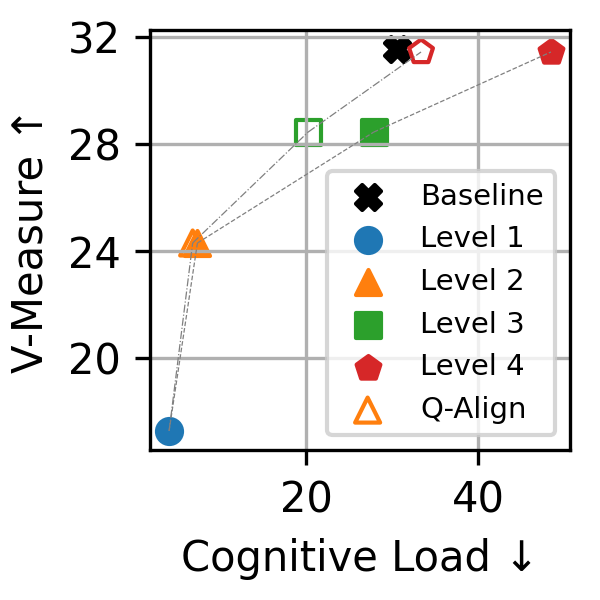}
        \subcaption{$\tau$ is 0.9}
    \end{minipage}
    \caption{Task performance vs. cognitive load on the text clustering task.}
    \label{fig:scores_and_cl_reduced_clustering}
\end{figure*}

\begin{figure*}[!ht]
    \centering
    \begin{minipage}[b]{0.25\linewidth}
        \centering
        \includegraphics[width=1.0\linewidth]{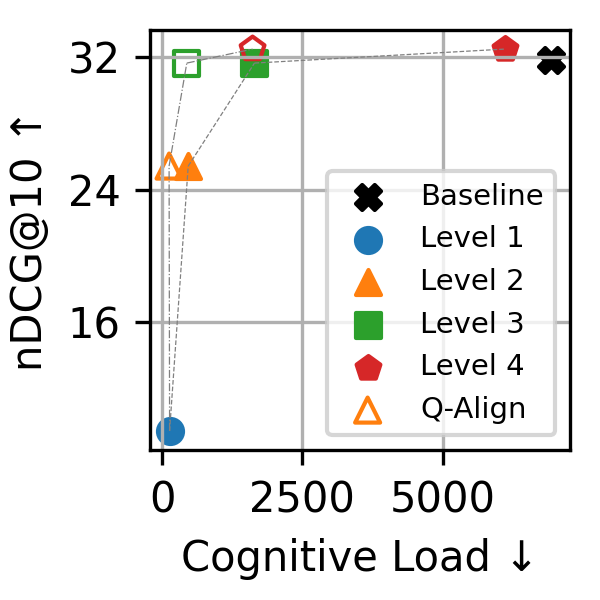}
        \subcaption{$\tau$ is 0.1}
    \end{minipage}
    \hspace{1cm}
    \begin{minipage}[b]{0.25\linewidth}
        \centering
        \includegraphics[width=1.0\linewidth]{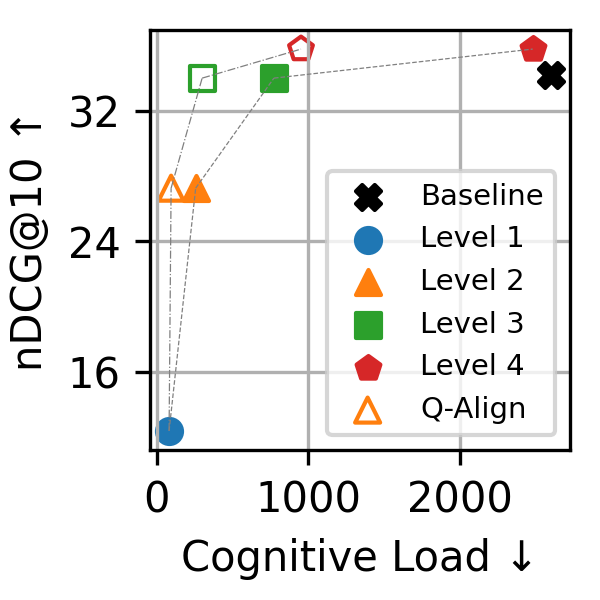}
        \subcaption{$\tau$ is 0.2}
    \end{minipage}
    \hspace{1cm}
    \begin{minipage}[b]{0.25\linewidth}
        \centering
        \includegraphics[width=1.0\linewidth]{figures/scores_and_cl_reduced/+2/with_t=False/retrieval/3.png}
        \subcaption{$\tau$ is 0.3}
    \end{minipage}
    \\\vspace{1cm}
    \begin{minipage}[b]{0.25\linewidth}
        \centering
        \includegraphics[width=1.0\linewidth]{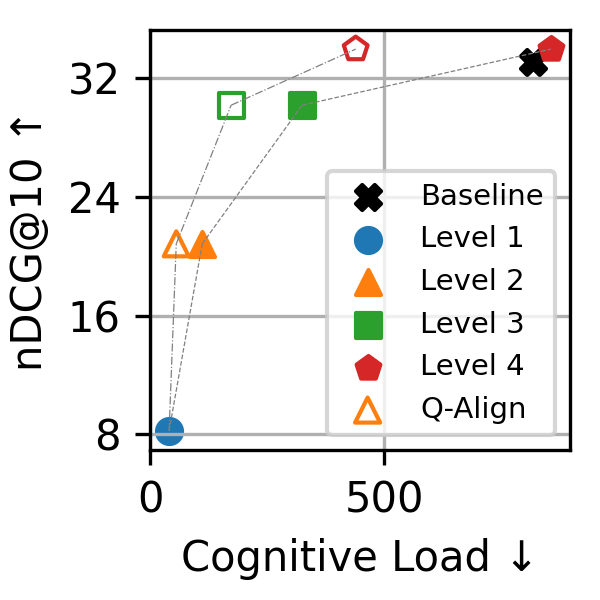}
        \subcaption{$\tau$ is 0.4}
    \end{minipage}
    \hspace{1cm}
    \begin{minipage}[b]{0.25\linewidth}
        \centering
        \includegraphics[width=1.0\linewidth]{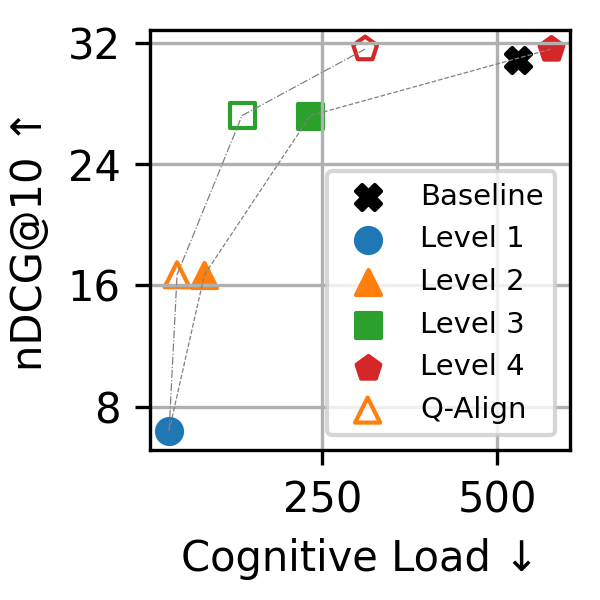}
        \subcaption{$\tau$ is 0.5}
    \end{minipage}
    \hspace{1cm}
    \begin{minipage}[b]{0.25\linewidth}
        \centering
        \includegraphics[width=1.0\linewidth]{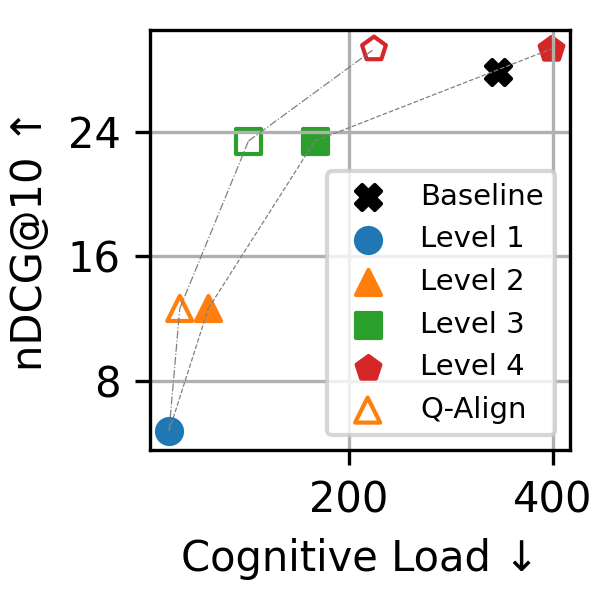}
        \subcaption{$\tau$ is 0.6}
    \end{minipage}
    \\\vspace{1cm}
    \begin{minipage}[b]{0.25\linewidth}
        \centering
        \includegraphics[width=1.0\linewidth]{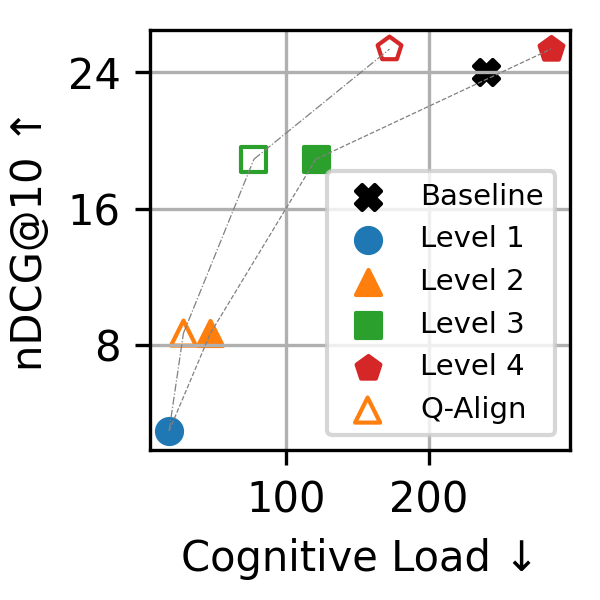}
        \subcaption{$\tau$ is 0.7}
    \end{minipage}
    \hspace{1cm}
    \begin{minipage}[b]{0.25\linewidth}
        \centering
        \includegraphics[width=1.0\linewidth]{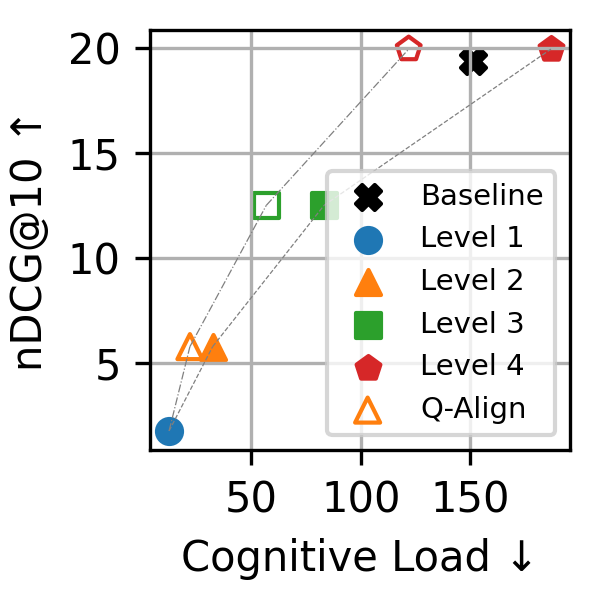}
        \subcaption{$\tau$ is 0.8}
    \end{minipage}
    \hspace{1cm}
    \begin{minipage}[b]{0.25\linewidth}
        \centering
        \includegraphics[width=1.0\linewidth]{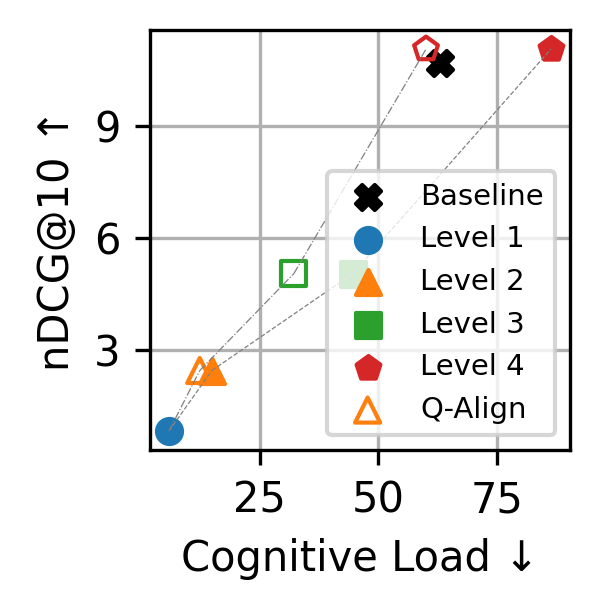}
        \subcaption{$\tau$ is 0.9}
    \end{minipage}
    \caption{Task performance vs. cognitive load on the information retrieval task.}
    \label{fig:scores_and_cl_reduced_retrieval}
\end{figure*}

Figures~\ref{fig:scores_and_cl_reduced_sts}--\ref{fig:scores_and_cl_reduced_retrieval} show the full results of the performance and cognitive load on the STS, text clustering, and information retrieval tasks, respectively.

\section{Task Performance by Dataset}
\label{apd:performance_by_dataset}
Tables~\ref{tab:performance_by_dataset_sts}--\ref{tab:performance_by_dataset_retrieval} present the performance on individual datasets for the STS, text clustering, and information retrieval tasks.
CQG-MBQA (Repro.) denotes our reimplementation of CQG-MBQA, which corresponds to the baseline described in the main text.
The results for all other methods, except CQG-MBQA (Repro.) and the proposed method, are taken from \citet{sun2025a}.
For CQG-MBQA (Ori.), CQG-MBQA (Repro.), and the proposed method, the threshold $\tau$ is set to 0.5.

\begin{table*}[!t]
\centering
\small
\begin{tabular}{clcccccccc}
    \toprule
    \multirow{2}{*}{\textbf{Type}} & \multirow{2}{*}{\textbf{Model}} & \multicolumn{8}{c}{\textbf{Spearman Correlation (STS)}} \\
    \cmidrule(lr){3-10}
    & & \textbf{STS12} & \textbf{STS13} & \textbf{STS14} & \textbf{STS15} & \textbf{STS16} & \textbf{STS-B} & \textbf{SICK-R} & \textbf{Avg.} \\
    \midrule
    \multirow{10}{*}{\rotatebox{90}{\textbf{Black-box}}}
    & \textbf{BERT} & 38.78 & 57.98 & 57.98 & 63.15 & 61.06 & 46.35 & 58.40 & 54.81 \\
    & \textbf{GloVe} & 54.64 & 69.16 & 60.81 & 72.31 & 65.34 & 61.54 & 55.43 & 62.74 \\
    & \textbf{USE} & 64.49 & 67.80 & 64.61 & 76.83 & 73.18 & 74.92 & 76.69 & 71.22 \\
    & \textbf{SimCSE (Unsup.)} & 66.05 & 81.49 & 73.61 & 79.72 & 78.12 & 76.52 & 72.24 & 75.39 \\
    & \textbf{SBERT (Ori.)} & 74.53 & 77.00 & 73.18 & 81.85 & 76.82 & 79.10 & 74.29 & 76.68 \\
    & \textbf{SimCSE (Sup.)} & 75.30 & 84.67 & 80.19 & 85.40 & 80.82 & 84.25 & 68.38 & 79.86 \\
    & \textbf{WhitenedCSE} & 74.65 & 85.79 & 77.49 & 84.71 & 80.33 & 81.48 & 75.34 & 79.97 \\
    & \textbf{SBERT (New)} & 73.08 & 82.13 & 76.73 & 85.58 & 80.23 & 83.09 & 79.32 & 80.02 \\
    & \textbf{OpenAI} & 72.84 & 86.1 & 81.15 & 88.49 & 85.08 & 83.56 & 79.00 & 82.31 \\
    & \textbf{AnglE} & 79.09 & 89.62 & 85.02 & 89.51 & 86.61 & 89.06 & 82.62 & 85.93 \\
    \midrule
    \multirow{8}{*}{\rotatebox{90}{\textbf{Interpretable}}}
    & \textbf{Bag-of-Tokens} & 44.75 & 52.06 & 54.78 & 68.65 & 60.59 & 54.85 & 57.87 & 56.22 \\
    & \textbf{QAEmb-MBQA} & 59.40 & 63.19 & 57.68 & 69.29 & 63.18 & 71.33 & 72.33 & 65.20 \\
    & \textbf{CQG-MBQA (Ori.)} & 69.21 & 80.19 & 73.91 & 80.66 & 78.30 & 82.69 & 78.21 & 77.60 \\
    & \textbf{CQG-MBQA (Repro.)} & 74.95 & 79.96 & 75.66 & 82.47 & 80.47 & 84.26 & 77.16 & 79.28 \\
    & \textbf{Ours (Level 4)} & 75.58 & 80.23 & 75.63 & 82.75 & 80.27 & 83.73 & 76.74 & 79.27 \\
    & \textbf{Ours (Level 3)} & 74.13 & 81.32 & 75.46 & 82.02 & 79.26 & 83.11 & 76.07 & 78.77 \\
    & \textbf{Ours (Level 2)} & 67.28 & 75.52 & 70.83 & 77.50 & 74.77 & 79.60 & 74.47 & 74.28 \\
    & \textbf{Ours (Level 1)} & 58.28 & 69.82 & 59.68 & 72.43 & 67.36 & 72.05 & 68.96 & 66.94 \\
    \bottomrule
\end{tabular}
\caption{Task performance by dataset on the STS task.}
\label{tab:performance_by_dataset_sts}
\end{table*}

\begin{table*}[!t]
\centering
\small
\begin{tabular}{clcccccccc}
    \toprule
    \multirow{2}{*}{\textbf{Type}} & \multirow{2}{*}{\textbf{Model}} & \multicolumn{8}{c}{\textbf{V-Measure (Clustering)}} \\
    \cmidrule(lr){3-10}
    & & \textbf{TNG} & \textbf{SE-P2P} & \textbf{BR-P2} & \textbf{BR-S2S} & \textbf{MR-P2P} & \textbf{MR-S2S} & \textbf{RD-P2P} & \textbf{Avg.} \\
    \midrule
    \multirow{7}{*}{\rotatebox{90}{\textbf{Black-box}}}
    & \textbf{SimCSE (Unsup.)} & 23.21 & 28.50 & 24.90 & 19.55 & 23.60 & 21.97 & 45.14 & 26.70 \\
    & \textbf{GloVe} & 25.83 & 28.51 & 29.27 & 19.18 & 26.12 & 20.38 & 35.82 & 26.44 \\
    & \textbf{BERT} & 23.35 & 26.55 & 30.12 & 24.77 & 26.09 & 23.60 & 43.32 & 28.26 \\
    & \textbf{SimCSE (Sup.)} & 34.86 & 29.45 & 30.15 & 24.67 & 26.25 & 24.12 & 47.74 & 31.03 \\
    & \textbf{SBERT (New)} & 47.47 & 33.13 & 36.99 & 33.21 & 34.25 & 32.24 & 54.80 & 38.87 \\
    & \textbf{AnglE} & 51.72 & 36.72 & 39.38 & 37.23 & 33.22 & 31.18 & 65.35 & 42.11 \\
    & \textbf{OpenAI} & 58.14 & 36.88 & 38.03 & 36.53 & 32.70 & 31.27 & 67.96 & 43.07 \\
    \midrule
    \multirow{8}{*}{\rotatebox{90}{\textbf{Interpretable}}}
    & \textbf{Bag-of-Tokens} & 8.52 & 17.64 & 4.70 & 3.32 & 11.39 & 13.05 & 15.67 & 10.61 \\
    & \textbf{QAEmb-MBQA} & 36.72 & 25.68 & 24.66 & 21.16 & 25.53 & 22.85 & 46.57 & 29.02 \\
    & \textbf{CQG-MBQA (Ori.)} & 40.00 & 28.22 & 34.88 & 31.13 & 31.02 & 28.71 & 54.40 & 35.48 \\
    & \textbf{CQG-MBQA (Repro.)} & 43.95 & 34.50 & 36.78 & 33.21 & 31.45 & 29.29 & 57.93 & 38.16 \\
    & \textbf{Ours (Level 4)} & 42.78 & 34.50 & 36.93 & 32.65 & 31.74 & 29.41 & 58.24 & 38.04 \\
    & \textbf{Ours (Level 3)} & 41.01 & 33.64 & 35.68 & 31.80 & 31.02 & 28.71 & 56.61 & 36.93 \\
    & \textbf{Ours (Level 2)} & 34.32 & 32.36 & 29.87 & 27.31 & 27.87 & 25.74 & 50.78 & 32.61 \\
    & \textbf{Ours (Level 1)} & 28.75 & 30.21 & 23.58 & 18.36 & 23.80 & 20.60 & 38.63 & 26.27 \\
    \bottomrule
\end{tabular}
\caption{Task performance by dataset on the text clustering task.}
\label{tab:performance_by_dataset_clustering}
\end{table*}

\begin{table*}[!t]
\centering
\small
\begin{tabular}{clccccccc}
    \toprule
    \multirow{2}{*}{\textbf{Type}} & \multirow{2}{*}{\textbf{Model}} & \multicolumn{7}{c}{\textbf{nDCG@10 (Retrieval)}} \\
    \cmidrule(lr){3-9}
    & & \textbf{MS MARCO} & \textbf{ArguAna} & \textbf{FQA} & \textbf{NFC} & \textbf{SCIDOCS} & \textbf{SciFact} & \textbf{Avg.} \\
    \midrule
    \multirow{7}{*}{\rotatebox{90}{\textbf{Black-box}}}
    & \textbf{BERT}                 & 16.86 & 28.29 & 2.19 & 4.30 & 2.82 & 13.34 & 11.30 \\
    & \textbf{SimCSE (Unsup.)}      & 44.63 & 38.34 & 9.84 & 9.88 & 5.50 & 25.72 & 22.32 \\
    & \textbf{GloVe}                & 44.27 & 36.30 & 10.09 & 13.87 & 8.04 & 29.58 & 23.69 \\
    & \textbf{SimCSE (Sup.)}        & 47.86 & 38.33 & 10.41 & 12.42 & 7.53 & 29.59 & 24.36 \\
    & \textbf{SBERT (New)}          & 88.74 & 47.13 & 37.27 & 32.25 & 21.82 & 62.64 & 48.31 \\
    & \textbf{AnglE}                & 90.43 & 66.15 & 44.84 & 38.65 & 22.98 & 74.07 & 56.19 \\
    & \textbf{OpenAI}               & 92.18 & 58.05 & 55.00 & 42.07 & 23.11 & 77.77 & 58.03 \\
    \midrule
    \multirow{9}{*}{\rotatebox{90}{\textbf{Interpretable}}}
    & \textbf{Bag-of-Tokens}        & 29.79 & 34.25 & 3.99 & 21.51 & 6.79 & 47.36 & 23.95 \\
    & \textbf{BM25}                 & 68.42 & 49.28 & 25.14 & 32.08 & 15.78 & 68.70 & 43.23 \\
    & \textbf{QAEmb-MBQA}           & 40.51 & 34.75 & 8.23 & 3.87 & 3.74 & 12.01 & 17.18 \\
    & \textbf{CQG-MBQA (Ori.)}      & 62.21 & 47.75 & 18.63 & 9.74 & 8.67 & 32.80 & 29.97 \\
    & \textbf{CQG-MBQA (Repro.)}    & 67.79 & 55.57 & 16.64 & 7.54 & 12.87 & 25.16 & 30.93 \\
    & \textbf{Ours (Level 4)}       & 69.66 & 55.04 & 18.63 & 8.36 & 12.65 & 25.40 & 31.62 \\
    & \textbf{Ours (Level 3)}       & 61.80 & 49.00 & 15.12 & 7.07 & 10.18 & 20.21 & 27.29 \\
    & \textbf{Ours (Level 2)}       & 38.29 & 35.64 & 5.62 & 5.47 & 5.69 & 9.54 & 16.71 \\
    & \textbf{Ours (Level 1)}       & 12.16 & 14.54 & 2.29 & 3.11 & 2.12 & 4.11 & 6.39 \\
    \bottomrule
\end{tabular}
\caption{Task performance by dataset on the information retrieval task.}
\label{tab:performance_by_dataset_retrieval}
\end{table*}

\section{Full Results of Analysis of Appropriate Granularity of Questions}
\label{apd:full_results_granularity_questions}
\begin{figure*}[!ht]
    \centering
    \begin{minipage}[b]{0.25\linewidth}
        \centering
        \includegraphics[width=1.0\linewidth]{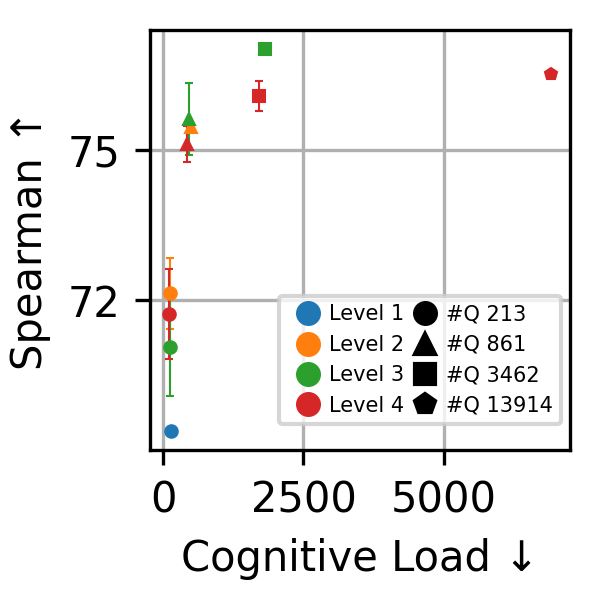}
        \subcaption{$\tau$ is 0.1}
    \end{minipage}
    \hspace{1cm}
    \begin{minipage}[b]{0.25\linewidth}
        \centering
        \includegraphics[width=1.0\linewidth]{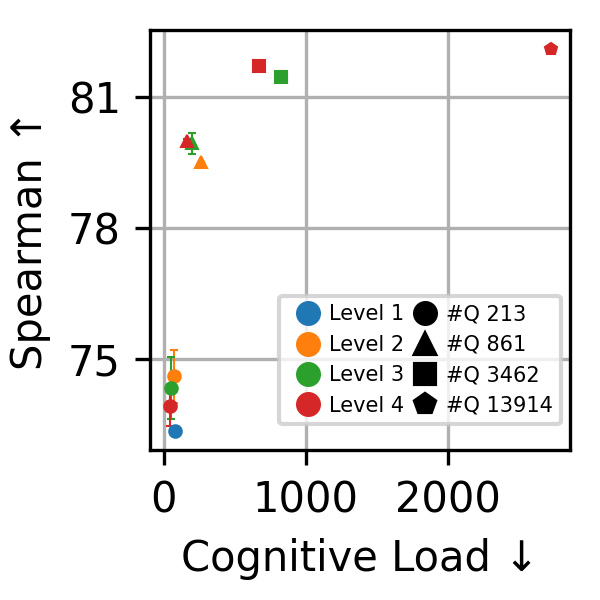}
        \subcaption{$\tau$ is 0.2}
    \end{minipage}
    \hspace{1cm}
    \begin{minipage}[b]{0.25\linewidth}
        \centering
        \includegraphics[width=1.0\linewidth]{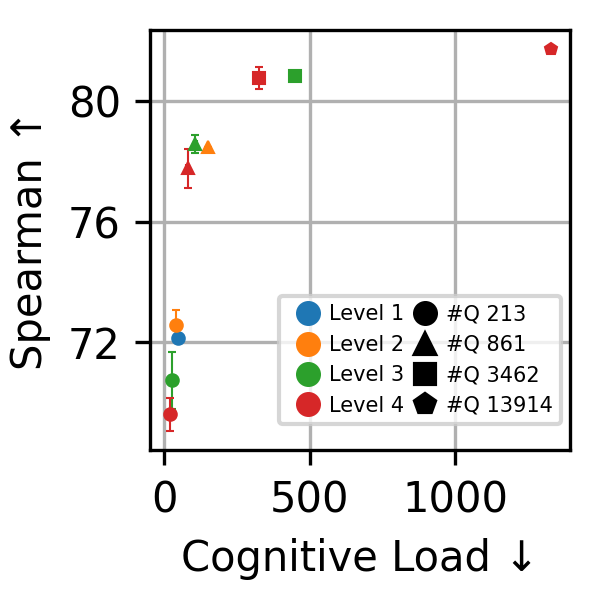}
        \subcaption{$\tau$ is 0.3}
    \end{minipage}
    \\\vspace{1cm}
    \begin{minipage}[b]{0.25\linewidth}
        \centering
        \includegraphics[width=1.0\linewidth]{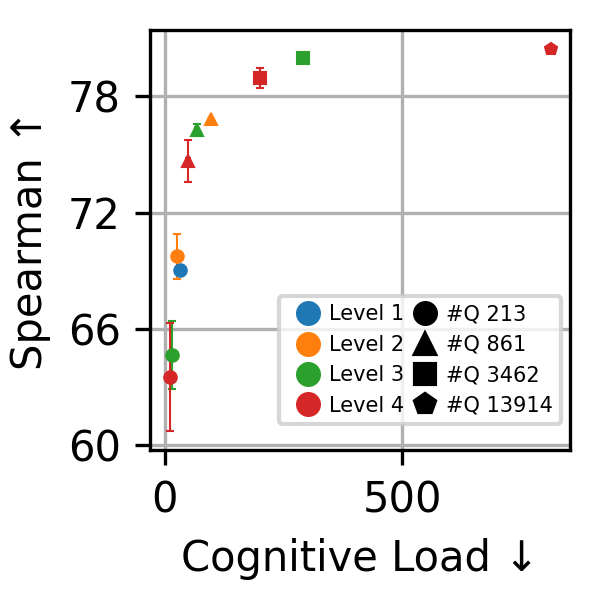}
        \subcaption{$\tau$ is 0.4}
    \end{minipage}
    \hspace{1cm}
    \begin{minipage}[b]{0.25\linewidth}
        \centering
        \includegraphics[width=1.0\linewidth]{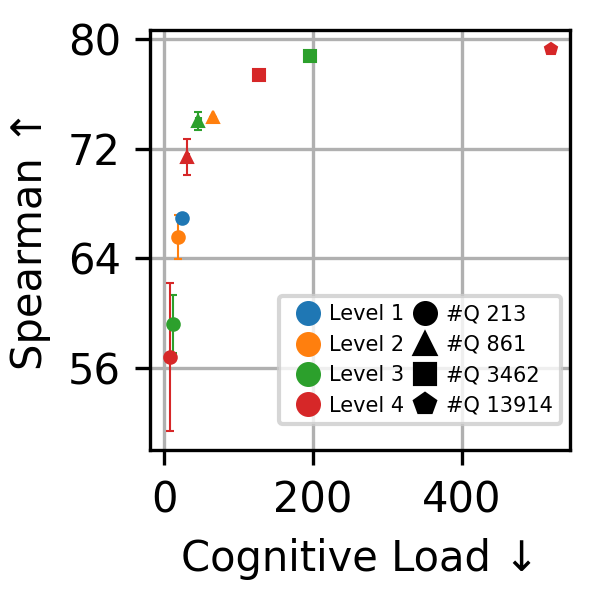}
        \subcaption{$\tau$ is 0.5}
    \end{minipage}
    \hspace{1cm}
    \begin{minipage}[b]{0.25\linewidth}
        \centering
        \includegraphics[width=1.0\linewidth]{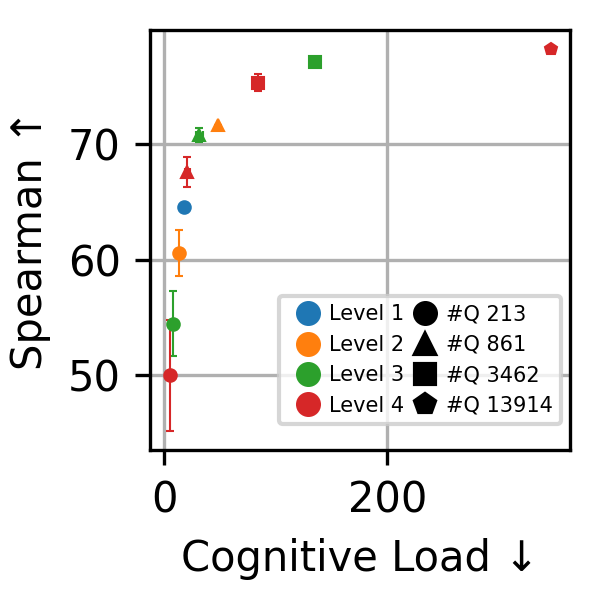}
        \subcaption{$\tau$ is 0.6}
    \end{minipage}
    \\\vspace{1cm}
    \begin{minipage}[b]{0.25\linewidth}
        \centering
        \includegraphics[width=1.0\linewidth]{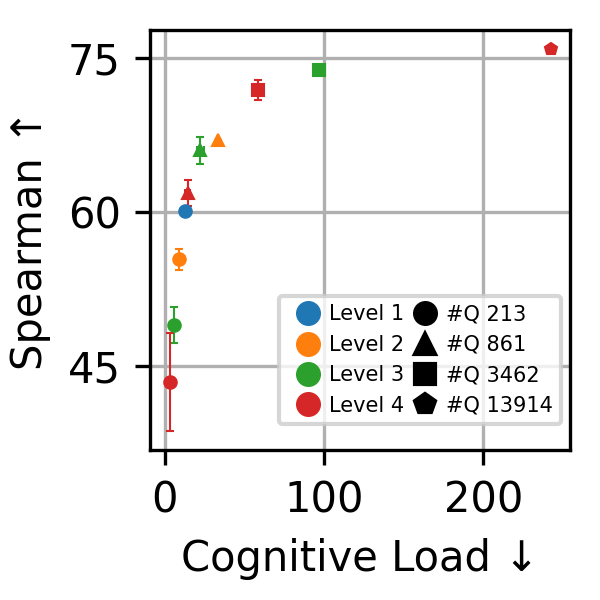}
        \subcaption{$\tau$ is 0.7}
    \end{minipage}
    \hspace{1cm}
    \begin{minipage}[b]{0.25\linewidth}
        \centering
        \includegraphics[width=1.0\linewidth]{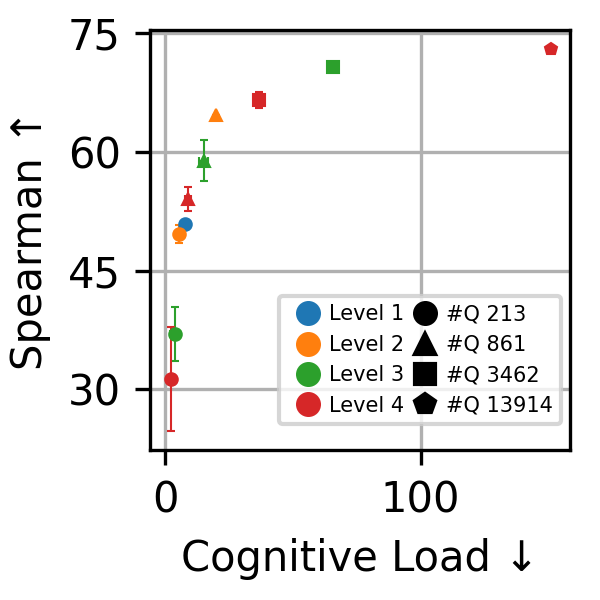}
        \subcaption{$\tau$ is 0.8}
    \end{minipage}
    \hspace{1cm}
    \begin{minipage}[b]{0.25\linewidth}
        \centering
        \includegraphics[width=1.0\linewidth]{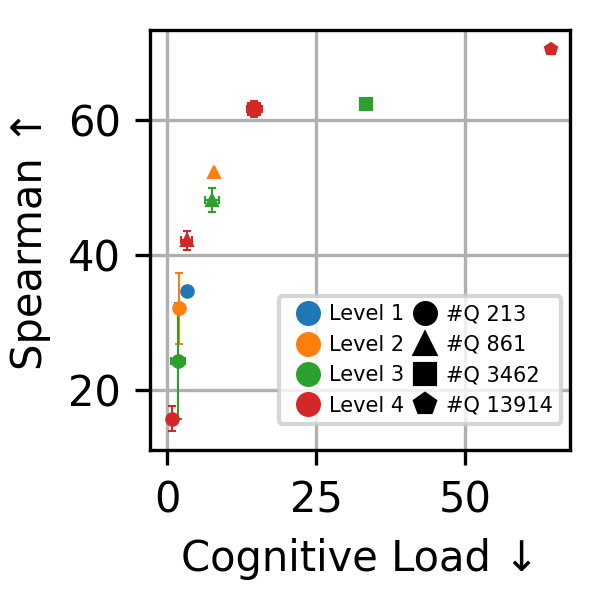}
        \subcaption{$\tau$ is 0.9}
    \end{minipage}
    \caption{Task performance vs. cognitive load on the STS task when the number of questions is matched across levels by random sampling.}
    \label{fig:scores_and_cl_random_sts}
\end{figure*}

\begin{figure*}[!ht]
    \centering
    \begin{minipage}[b]{0.25\linewidth}
        \centering
        \includegraphics[width=1.0\linewidth]{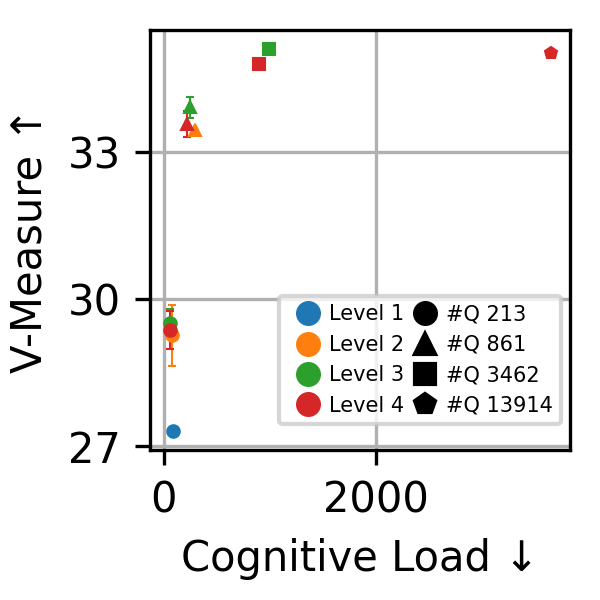}
        \subcaption{$\tau$ is 0.1}
    \end{minipage}
    \hspace{1cm}
    \begin{minipage}[b]{0.25\linewidth}
        \centering
        \includegraphics[width=1.0\linewidth]{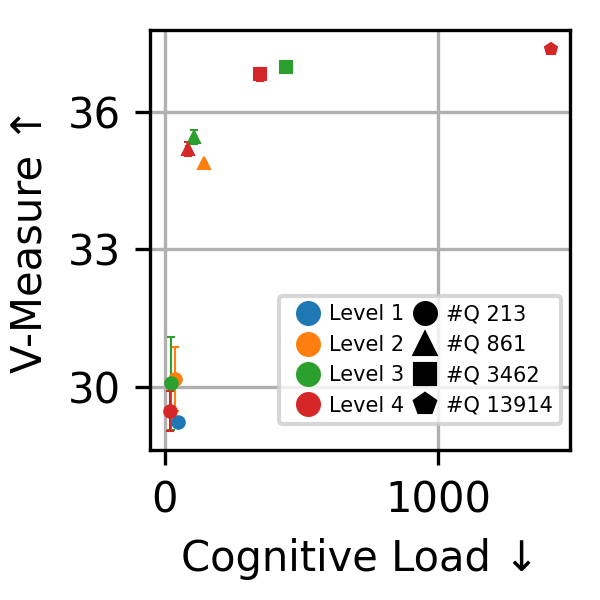}
        \subcaption{$\tau$ is 0.2}
    \end{minipage}
    \hspace{1cm}
    \begin{minipage}[b]{0.25\linewidth}
        \centering
        \includegraphics[width=1.0\linewidth]{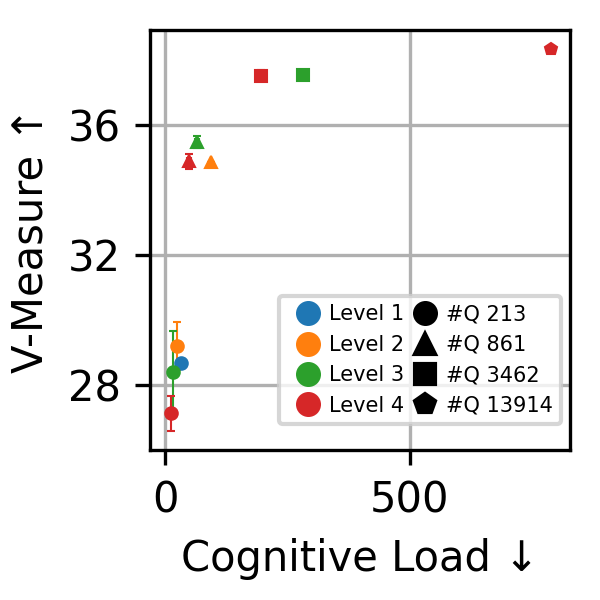}
        \subcaption{$\tau$ is 0.3}
    \end{minipage}
    \\\vspace{1cm}
    \begin{minipage}[b]{0.25\linewidth}
        \centering
        \includegraphics[width=1.0\linewidth]{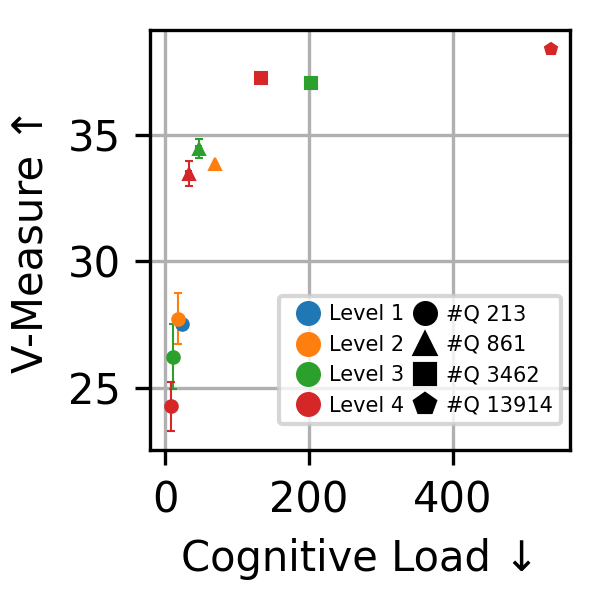}
        \subcaption{$\tau$ is 0.4}
    \end{minipage}
    \hspace{1cm}
    \begin{minipage}[b]{0.25\linewidth}
        \centering
        \includegraphics[width=1.0\linewidth]{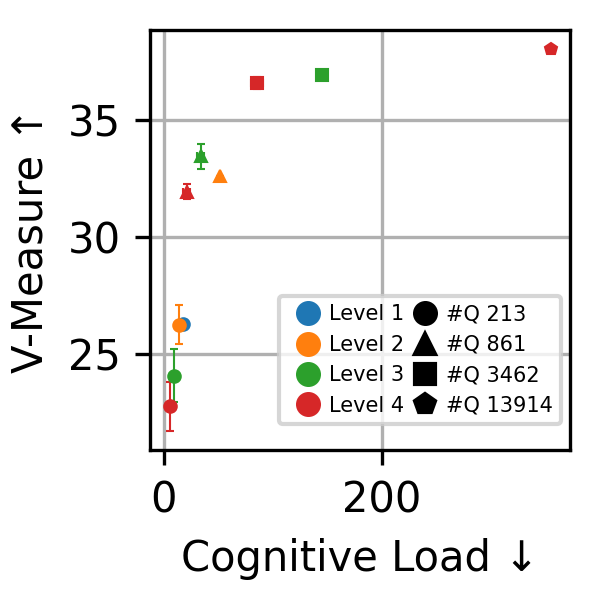}
        \subcaption{$\tau$ is 0.5}
    \end{minipage}
    \hspace{1cm}
    \begin{minipage}[b]{0.25\linewidth}
        \centering
        \includegraphics[width=1.0\linewidth]{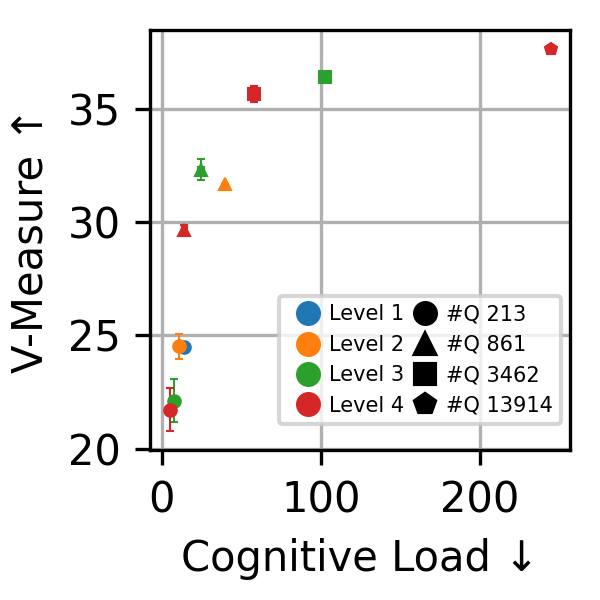}
        \subcaption{$\tau$ is 0.6}
    \end{minipage}
    \\\vspace{1cm}
    \begin{minipage}[b]{0.25\linewidth}
        \centering
        \includegraphics[width=1.0\linewidth]{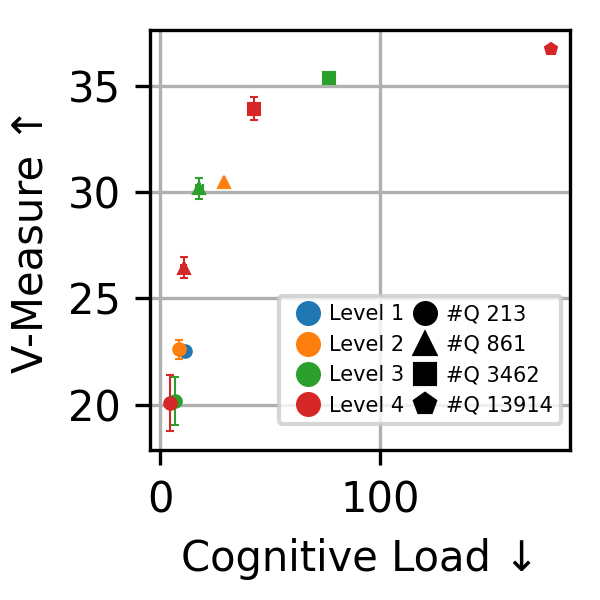}
        \subcaption{$\tau$ is 0.7}
    \end{minipage}
    \hspace{1cm}
    \begin{minipage}[b]{0.25\linewidth}
        \centering
        \includegraphics[width=1.0\linewidth]{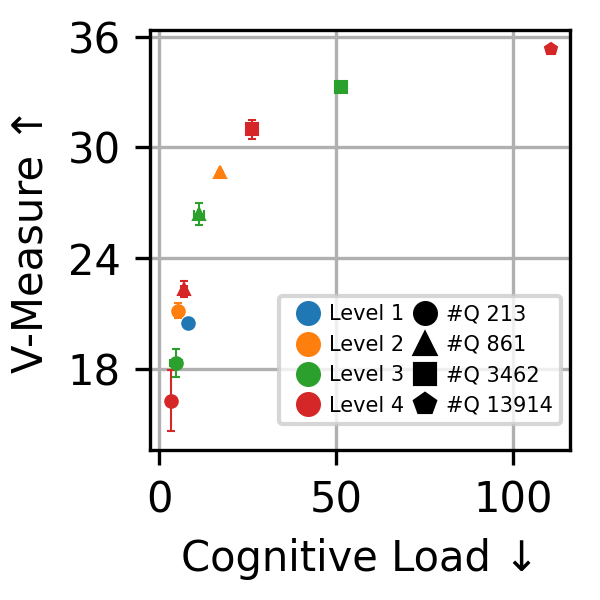}
        \subcaption{$\tau$ is 0.8}
    \end{minipage}
    \hspace{1cm}
    \begin{minipage}[b]{0.25\linewidth}
        \centering
        \includegraphics[width=1.0\linewidth]{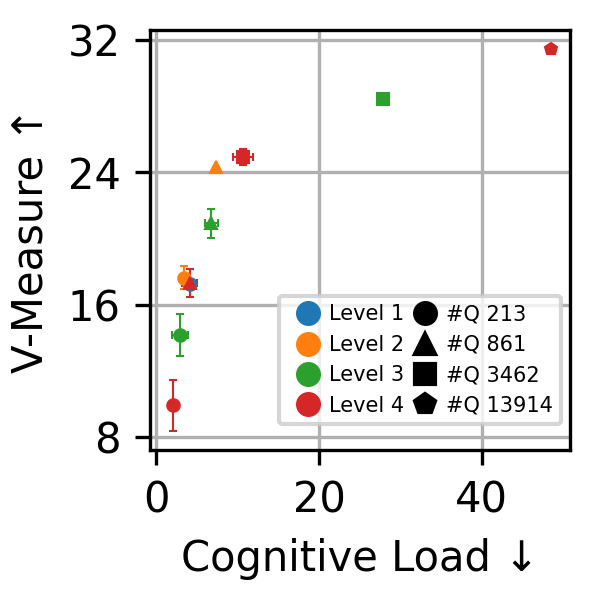}
        \subcaption{$\tau$ is 0.9}
    \end{minipage}
    \caption{Task performance vs. cognitive load on the text clustering task when the number of questions is matched across levels by random sampling.}
    \label{fig:scores_and_cl_random_clustering}
\end{figure*}

\begin{figure*}[!ht]
    \centering
    \begin{minipage}[b]{0.25\linewidth}
        \centering
        \includegraphics[width=1.0\linewidth]{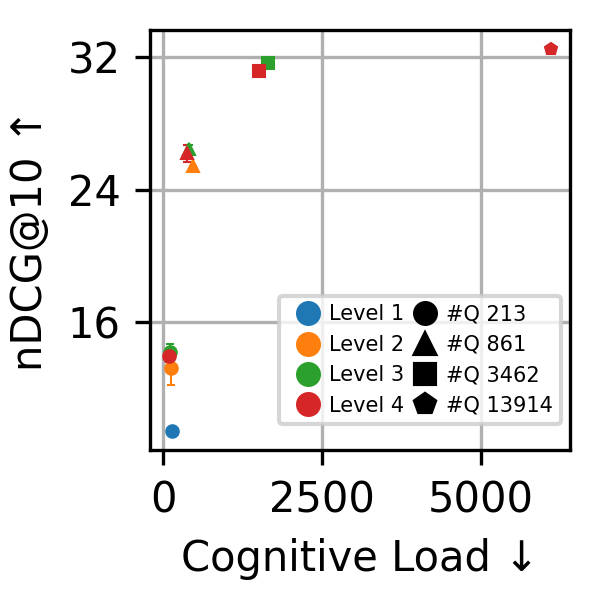}
        \subcaption{$\tau$ is 0.1}
    \end{minipage}
    \hspace{1cm}
    \begin{minipage}[b]{0.25\linewidth}
        \centering
        \includegraphics[width=1.0\linewidth]{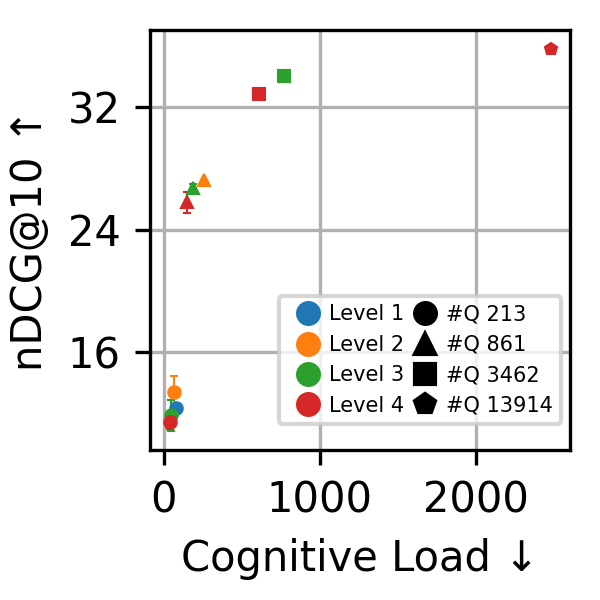}
        \subcaption{$\tau$ is 0.2}
    \end{minipage}
    \hspace{1cm}
    \begin{minipage}[b]{0.25\linewidth}
        \centering
        \includegraphics[width=1.0\linewidth]{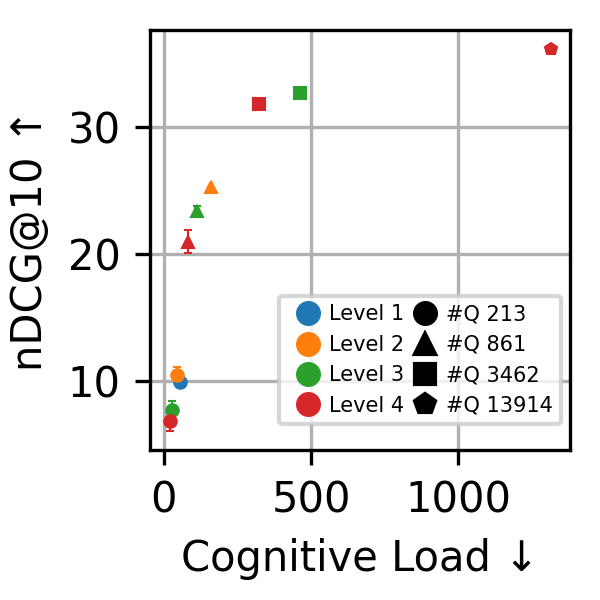}
        \subcaption{$\tau$ is 0.3}
    \end{minipage}
    \\\vspace{1cm}
    \begin{minipage}[b]{0.25\linewidth}
        \centering
        \includegraphics[width=1.0\linewidth]{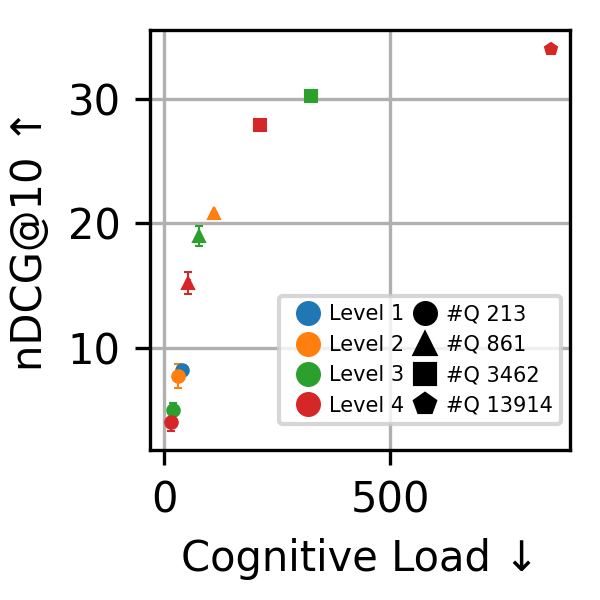}
        \subcaption{$\tau$ is 0.4}
    \end{minipage}
    \hspace{1cm}
    \begin{minipage}[b]{0.25\linewidth}
        \centering
        \includegraphics[width=1.0\linewidth]{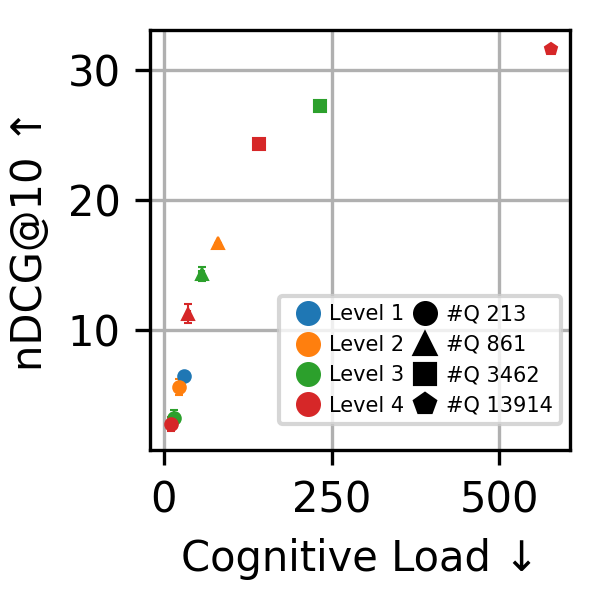}
        \subcaption{$\tau$ is 0.5}
    \end{minipage}
    \hspace{1cm}
    \begin{minipage}[b]{0.25\linewidth}
        \centering
        \includegraphics[width=1.0\linewidth]{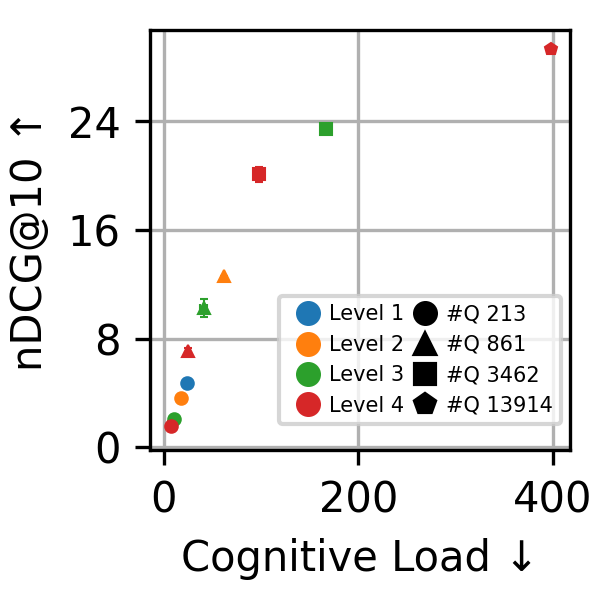}
        \subcaption{$\tau$ is 0.6}
    \end{minipage}
    \\\vspace{1cm}
    \begin{minipage}[b]{0.25\linewidth}
        \centering
        \includegraphics[width=1.0\linewidth]{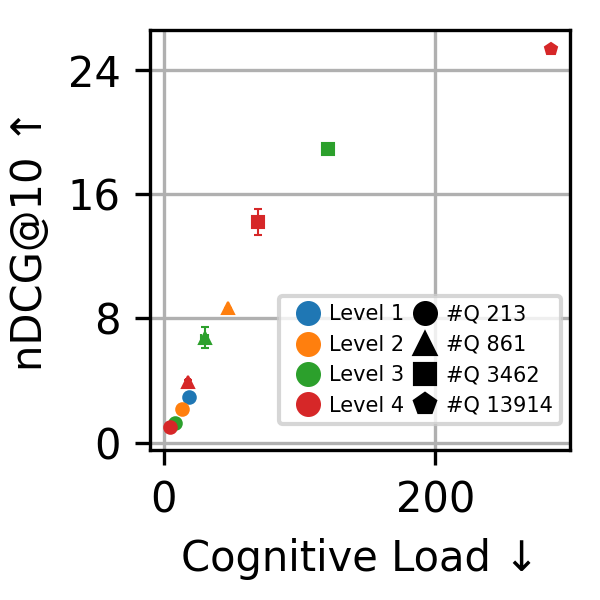}
        \subcaption{$\tau$ is 0.7}
    \end{minipage}
    \hspace{1cm}
    \begin{minipage}[b]{0.25\linewidth}
        \centering
        \includegraphics[width=1.0\linewidth]{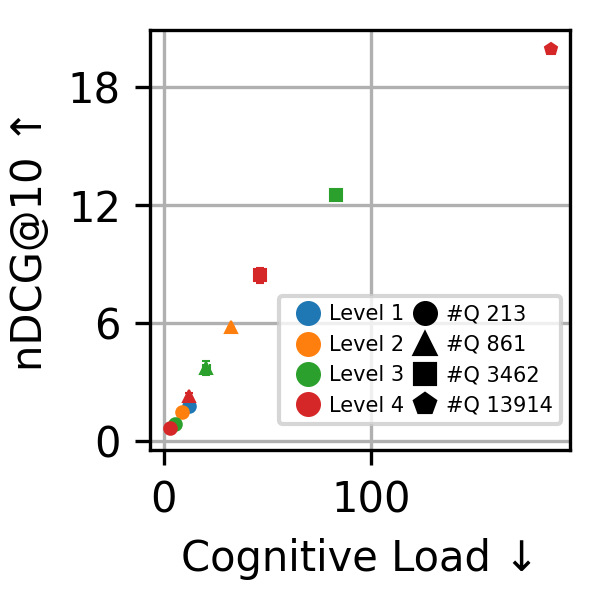}
        \subcaption{$\tau$ is 0.8}
    \end{minipage}
    \hspace{1cm}
    \begin{minipage}[b]{0.25\linewidth}
        \centering
        \includegraphics[width=1.0\linewidth]{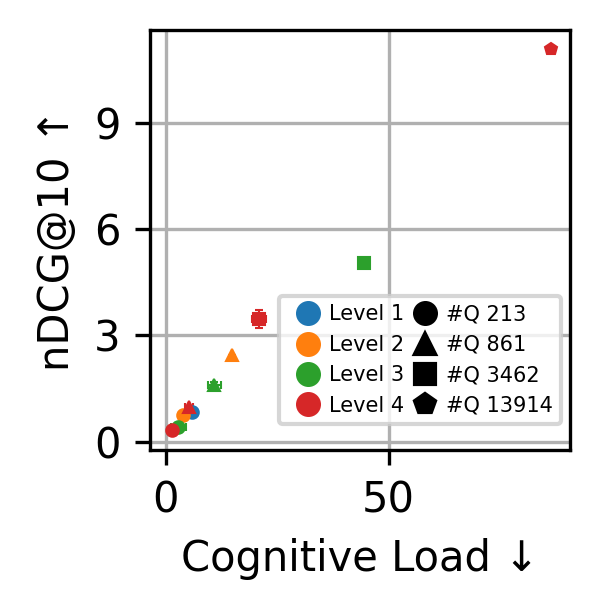}
        \subcaption{$\tau$ is 0.9}
    \end{minipage}
    \caption{Task performance vs. cognitive load on the information retrieval task when the number of questions is matched across levels by random sampling.}
    \label{fig:scores_and_cl_random_retrieval}
\end{figure*}

Figures~\ref{fig:scores_and_cl_random_sts}--\ref{fig:scores_and_cl_random_retrieval} show the full results of the performance and cognitive load on the STS, text clustering, and information retrieval tasks when the number of questions is matched across levels by random sampling, respectively.

\end{document}